\documentclass{article}

\usepackage[preprint,nonatbib]{neurips_2021}

\usepackage[utf8]{inputenc} 
\usepackage[T1]{fontenc}    
\usepackage{hyperref}       
\hypersetup{colorlinks,citecolor=cyan,linkcolor=magenta}
\usepackage{url}            
\usepackage{booktabs}       
\usepackage{amsfonts}       
\usepackage{nicefrac}       
\usepackage{microtype}      
\usepackage{xcolor}         
\usepackage{xspace}
\usepackage{multirow}
\usepackage{enumitem}
\usepackage{caption}
\usepackage{subcaption}
\usepackage{xfrac}
\usepackage{makecell}
\usepackage{float}

\usepackage{amsmath,amsfonts,bm}

\def\rvx{{\mathbf{x}}}

\def\rmP{{\mathbf{P}}}

\DeclareMathAlphabet{\mathsfit}{\encodingdefault}{\sfdefault}{m}{sl}
\SetMathAlphabet{\mathsfit}{bold}{\encodingdefault}{\sfdefault}{bx}{n}

\definecolor{airforceblue}{rgb}{0.36, 0.54, 0.66}
\definecolor{ao(english)}{rgb}{0.0, 0.7, 0.0}
\definecolor{azure(colorwheel)}{rgb}{0.0, 0.5, 1.0}

\usepackage{mathtools}
\DeclarePairedDelimiter\roundbracket{(}{)}
\DeclarePairedDelimiter\squarebracket{[}{]}
\DeclarePairedDelimiter\curlybracket{\{}{\}}
\makeatletter
\def\rbr{\@ifnextchar[{\roundbracket}{\roundbracket*}}
\def\sbr{\@ifnextchar[{\squarebracket}{\squarebracket*}}
\def\cbr{\@ifnextchar[{\curlybracket}{\curlybracket*}}
\makeatother

\newcommand{\name}{CoAtNet\xspace}

\newcommand{\expo}[1]{\exp \left( {#1} \right)}

\newcommand{\mc}[1]{\mathcal{#1}}

\newcommand{\mbb}[1]{\mathbb{#1}}

\title{
\name: Marrying Convolution and Attention \\ for All Data Sizes
}

\author{%
  Zihang Dai, 
  Hanxiao Liu,
  Quoc V. Le, 
  Mingxing Tan \\
  Google Research, Brain Team \\
  \texttt{\{zihangd,hanxiaol,qvl,tanmingxing\}@google.com} \\
}

\begin{document}

\maketitle

\begin{abstract}
  Transformers have attracted increasing interests in computer vision, but they still fall behind state-of-the-art convolutional networks. In this work, we show that while Transformers tend to have larger model capacity, their generalization can be worse than convolutional networks due to the lack of the right inductive bias. To effectively combine the strengths from both architectures, we present \name{s} (pronounced ``coat'' nets), a family of hybrid models built from two key insights: (1) depthwise \textbf{\underline{Co}}nvolution and self-\textbf{\underline{At}}tention can be naturally unified via simple relative attention; (2) vertically stacking convolution layers and attention layers in a principled way is surprisingly effective in improving generalization, capacity and efficiency.
  Experiments show that our \name{s} achieve state-of-the-art performance under different resource constraints across various datasets:
  Without extra data, \name achieves 86.0\% ImageNet top-1 accuracy;
  When pre-trained with 13M images from ImageNet-21K, our \name achieves 88.56\% top-1 accuracy, matching ViT-huge pre-trained with 300M images from JFT-300M while using 23x less data;
  Notably, when we further scale up \name with  JFT-3B, it achieves 90.88\%  top-1 accuracy on ImageNet, establishing a new state-of-the-art result.
\end{abstract}

\section{Introduction}
\label{sec:intro}
Since the breakthrough of AlexNet~\cite{alexnet12}, Convolutional Neural Networks (ConvNets) have been the dominating model architecture for computer vision~\cite{simonyan2014very, he2016deep, szegedy2015going, efficientnet19}. 
Meanwhile, with the success of self-attention models like Transformers~\cite{vaswani2017attention} in natural language processing~\cite{devlin2018bert, brown2020language}, many previous works have attempted to bring in the power of attention into computer vision~\cite{wang2018non,bello2019attention,srinivas2021bottleneck,shen2021efficient}.
More recently, Vision Transformer (ViT)~\cite{dosovitskiy2020image} has shown that with almost\footnote{The initial projection stage can be seen as an aggressive down-sampling convolutional stem.} only vanilla Transformer layers, one could obtain reasonable performance on ImageNet-1K~\cite{deng2009imagenet} alone. More importantly, when pre-trained on large-scale weakly labeled JFT-300M dataset~\cite{sun2017revisiting}, ViT achieves comparable results to state-of-the-art (SOTA) ConvNets, indicating that Transformer models potentially have higher capacity at scale than ConvNets.

While ViT has shown impressive results with enormous JFT 300M training images,  its performance still falls behind ConvNets in the low data regime. For example, without extra JFT-300M pre-training, the ImageNet accuracy of ViT is still significantly lower than ConvNets with comparable model size~\cite{efficientnet19} (see Table \ref{tab:i1k_result}).
Subsequent works use special regularization and stronger data augmentation to improve the vanilla ViT~\cite{touvron2020training,touvron2021going,zhou2021deepvit},
yet none of these ViT variants could outperform the SOTA \textit{convolution-only} models on ImageNet classification given the same amount of data and computation~\cite{tan2021efficientnetv2,brock2021high}.
This suggests that vanilla Transformer layers may lack certain desirable inductive biases possessed by ConvNets, and thus require significant amount of data and computational resource to compensate. Not surprisingly, many recent works have been trying to incorporate the inductive biases of ConvNets into Transformer models, by imposing local receptive fields for attention layers~\cite{vaswani2021scaling,liu2021swin} or 
augmenting the attention and FFN layers with implicit or explicit convolutional operations~\cite{wu2021cvt,graham2021levit,yuan2021tokens}.
However, these approaches are either ad-hoc or focused on injecting a particular property, lacking a systematic understanding of the respective roles of convolution and attention when combined.

In this work, we systematically study the problem of hybridizing convolution and attention from two fundamental aspects in machine learning -- generalization and model capacity.
Our study shows that convolutional layers tend to have better generalization with faster converging speed thanks to their strong prior of inductive bias, while attention layers have higher model capacity that can benefit from larger datasets.
Combining convolutional and attention layers can achieve better generalization and capacity; however, a key challenge here is how to effectively combine them to achieve better trade-offs between accuracy and efficiency.
In this paper, we investigate two key insights: First, we observe that the commonly used depthwise convolution can be effectively merged into attention layers with simple relative attention; Second, simply stacking convolutional and attention layers, in a proper way, could be surprisingly effective to achieve better generalization and capacity.
Based on these insights, we propose a simple yet effective network architecture named \name, which enjoys the strengths from both ConvNets and Transformers.

Our \name achieves SOTA performances under comparable resource constraints across different data sizes.
Specifically, under the low-data regime, \name inherits the great generalization property of ConvNets thanks to the favorable inductive biases.
Moreover, given abundant data, \name not only enjoys the superior scalability of Transformer models, but also achieves faster convergence and thus improved efficiency.
When only ImageNet-1K is used for training, \name achieves 86.0\% top-1 accuracy, matching the prior art NFNet~\cite{brock2021high} under similar computation resource and training conditions.
Further, when pre-trained on ImageNet-21K with about 10M images, \name reaches 88.56\% top-1 accuracy when finetuned on ImageNet-1K, matching the ViT-Huge pre-trained on JFT-300M, a 23$\times$ larger dataset.
Finally, when JFT-3B is used for pre-training, \name exhibits better efficiency compared to ViT, and pushes the ImageNet-1K top-1 accuracy to 90.88\% while using 1.5x less computation of the prior art set by ViT-G/14~\cite{zhai2021scaling}.

\section{Model}
\label{sec:model}
In the section, we focus on the question of how to ``optimally'' combine the convolution and transformer.
Roughly speaking, we decompose the question into two parts:
\begin{enumerate}[itemsep=0pt,topsep=0pt,partopsep=0pt]
    \item How to combine the convolution and self-attention within one basic computational block?
    \item How to vertically stack different types of computational blocks together to form a complete network?
\end{enumerate}
The rationale of the decomposition will become clearer as we gradually reveal our design choices.

\subsection{Merging Convolution and Self-Attention}
For convolution, we mainly focus on the MBConv block~\cite{sandler2018mobilenetv2} which employs depthwise convolution~\cite{sepconv14} to capture the spatial interaction.
A key reason of this choice is that both the FFN module in Transformer and MBConv employ the design of ``inverted bottleneck'', which first expands the channel size of the input by 4x and later project the 
the 4x-wide hidden state back to the original channel size to enable residual connection.

Besides the similarity of inverted bottleneck, we also notice that both depthwise convolution and self-attention can be expressed as a per-dimension weighted sum of values in a pre-defined receptive field.
Specifically, convolution relies on a fixed kernel to gather information from a local receptive field
\begin{equation}
\label{eqn:conv}
y_i = \sum_{j \in \mathcal{L}(i)} w_{i-j} \odot x_j \quad \text{(depthwise convolution)},
\end{equation}
where $x_i, y_i \in \mbb{R}^D$ are the input and output at position $i$ respectively, and $\mathcal{L}(i)$ denotes a local neighborhood of $i$, e.g., a 3x3 grid centered at $i$ in image processing.

In comparison, self-attention allows the receptive field to be the entire spatial locations and computes the weights based on the re-normalized pairwise similarity between the pair $(x_i, x_j)$:\footnote{To simplify the presentation, we deliberately omit the multi-head query, key and value projections for now. In the actual implementation, we always use the multi-head projections.}
\begin{equation}
\label{eqn:attn}
    y_i = \sum_{j \in \mathcal{G}} \underbrace{\frac{\expo{x_i^\top x_j}}{\sum_{k \in \mathcal{G}} \expo{x_i^\top x_k}}}_{A_{i,j}} x_j \quad \text{(self-attention)},
\end{equation}
where $\mathcal{G}$ indicates the global spatial space.
Before getting into the question of how to best combine them, it is worthwhile to compare their relative strengths and weaknesses, which helps to figure out the good properties we hope to retain.
\begin{itemize}[leftmargin=*,itemsep=0pt,topsep=0pt]
\item First of all, the depthwise convolution kernel $w_{i-j}$ is an input-independent parameter of static value, while the attention weight $A_{i,j}$ dynamically depends on the representation of the input.
Hence, it is much easier for the self-attention to capture complicated relational interactions between different spatial positions, a property that we desire most when processing high-level concepts.
However, the flexibility comes with a risk of easier overfitting, especially when data is limited.

\item Secondly, notice that given any position pair $(i, j)$, the corresponding convolution weight $w_{i-j}$ only cares about the relative shift between them, i.e. $i - j$, rather than the specific values of $i$ or $j$.
This property is often referred to translation equivalence, which has been found to improve generalization under datasets of limited size~\cite{mohamed2020data}.
Due to the usage of absolution positional embeddings, standard Transformer (ViT) lacks this property.
This partially explains why ConvNets are usually better than Transformers when the dataset is not enormously large.

\item 
Finally, the size of the receptive field is one of the most crucial differences between self-attention and convolution.
Generally speaking, a larger receptive field provides more contextual information, which could lead to higher model capacity.
Hence, the global receptive field has been a key motivation to employ self-attention in vision.
However, a large receptive field requires significantly more computation.
In the case of global attention, the complexity is quadratic w.r.t. spatial size, which has been a fundamental trade-off in applying self-attention models.
\end{itemize}
\begin{table}[!ht]
    \centering
    \caption{Desirable properties found in convolution or self-attention.}
    \begin{tabular}{c|c|c}
        \toprule
        \bf Properties & \bf Convolution & \bf Self-Attention  \\
        \midrule
        Translation Equivariance & $\checkmark$ & \\
        Input-adaptive Weighting & & $\checkmark$ \\
        Global Receptive Field   & & $\checkmark$ \\
        \bottomrule
    \end{tabular}
    \label{tab:property_comparison}
\end{table}
Given the comparison above, an ideal model should be able to combine the 3 desirable properties in Table \ref{tab:property_comparison}.
With the similar form of depthwise convolution in Eqn. \eqref{eqn:conv} and self-attention in Eqn. \eqref{eqn:attn}, a straightforward idea that could achieve this is simply to sum a \textit{global} static convolution kernel with the adaptive attention matrix, either after or before the Softmax normalization, i.e.,
\begin{align}
\label{eqn:rel_attn}
    y_i^\text{post} = \sum_{j \in \mathcal{G}} \left(\frac{\expo{x_i^\top x_j}}{\sum_{k \in \mathcal{G}} \expo{x_i^\top x_k}} + w_{i-j} \right) x_j 
    \;\;\text{or}\;\;
    y_i^\text{pre} = \sum_{j \in \mathcal{G}} \frac{\expo{x_i^\top x_j + w_{i-j}}}{\sum_{k \in \mathcal{G}} \expo{x_i^\top x_k + w_{i-k}}} x_j.
\end{align}
Interestingly, while the idea seems overly simplified, the pre-normalization version $y^\text{pre}$ corresponds to a particular variant of relative self-attention~\cite{shaw2018self,raffel2019exploring}.
In this case, the attention weight $A_{i,j}$ is decided jointly by the $w_{i-j}$ of translation equivariance and the input-adaptive $x_i^\top x_j$, which can enjoy both effects depending on their relative magnitudes.
Importantly, note that in order to enable the global convolution kernel without blowing up the number of parameters, we have reloaded the notation of $w_{i-j}$ as a scalar (i.e., $w \in \mbb{R}^{O(|\mc{G}|)}$) rather than a vector in Eqn. \eqref{eqn:conv}.
Another advantage of the scalar formulation of $w$ is that retrieving $w_{i-j}$ for all $(i, j)$ is clearly subsumed by computing the pairwise dot-product attention, hence resulting in minimum additional cost (see Appendix \ref{sec:appendix_model_detail}).
Given the benefits, we will use the Transformer block with the \textit{pre-normalization} relative attention variant in Eqn. \eqref{eqn:rel_attn} as the key component of the proposed \name model.

\subsection{Vertical Layout Design}
After figuring out a neat way to combine convolution and attention, we next consider how to utilize it to stack an entire network.

As we have discuss above, the global context has a quadratic complexity w.r.t. the spatial size.
Hence, if we directly apply the relative attention in Eqn. \eqref{eqn:rel_attn} to the raw image input, the computation will be excessively slow due to the large number of pixels in any image of common sizes.
Hence, to construct a network that is feasible in practice, we have mainly three options:
\begin{itemize}[leftmargin=2em,itemsep=0pt,topsep=0pt,partopsep=0pt]
\item[(A)] Perform some down-sampling to reduce the spatial size and employ the global relative attention after the feature map reaches manageable level.
\item[(B)] Enforce local attention, which restricts the global receptive field $\mathcal{G}$ in attention to a local field $\mathcal{L}$ just like in convolution~\cite{liu2021swin,vaswani2021scaling}.
\item[(C)] Replace the quadratic Softmax attention with certain linear attention variant which only has a linear complexity w.r.t. the spatial size~\cite{shen2021efficient,katharopoulos2020transformers,choromanski2020rethinking}.
\end{itemize}
We briefly experimented with option (C) without getting a reasonably good result.
For option (B), we found that implementing local attention involves many non-trivial shape formatting operations that requires intensive memory access. 
On our accelerator of choice (TPU), such operation turns out to be extremely slow~\cite{ramachandran2019stand}, which not only defeats the original purpose of speeding up global attention, but also hurts the model capacity.
Hence, as some recent work has studied this variant~\cite{liu2021swin,vaswani2021scaling}, we will focus on option (A) and compare our results with theirs in our empirical study (Section \ref{sec:experiments}).

For option (A), the down-sampling can be achieved by either (1) a convolution stem with aggressive stride (e.g., stride 16x16) as in ViT or (2) a multi-stage network with gradual pooling as in ConvNets.
With these choices, we derive a search space of 5 variants and compare them in controlled experiments.
\begin{itemize}[leftmargin=*,itemsep=0pt,topsep=0pt,partopsep=0pt]
\item When the ViT Stem is used, we directly stack $L$ Transformer blocks with relative attention, which we denote as \textsc{ViT$_\text{rel}$}.

\item When the multi-stage layout is used, we mimic ConvNets to construct a network of 5 stages (\texttt{S0}, \texttt{S1}, \texttt{S2}, \texttt{S3} \& \texttt{S4}), with spatial resolution gradually decreased from \texttt{S0} to \texttt{S4}.  At the beginning of each stage, we always reduce the spatial size by 2x and increase the number of channels (see Appendix~\ref{sec:appendix_model_detail} for the detailed down-sampling implementation). 

The first stage \texttt{S0} is a simple 2-layer convolutional Stem and \texttt{S1} always employs MBConv blocks with squeeze-excitation (SE), as the spatial size is too large for global attention. 
Starting from \texttt{S2} through \texttt{S4}, we consider either the MBConv or the Transformer block, with a constraint that convolution stages must appear before Transformer stages.
The constraint is based on the prior that convolution is better at processing local patterns that are more common in early stages.
This leads to 4 variants with increasingly more Transformer stages, \textsc{C-C-C-C}, \textsc{C-C-C-T}, \textsc{C-C-T-T} and \textsc{C-T-T-T}, where \textsc{C} and \textsc{T} denote \underline{C}onvolution and \underline{T}ransformer respectively.

\end{itemize}

To systematically study the design choices, we consider two fundamental aspects generalization capability and model capacity: For \textbf{generalization}, we are interested in the gap between the training loss and the evaluation accuracy. If two models have the same training loss, then the model with higher evaluation accuracy has better generalization capability, since it can generalize better to unseen evaluation dataset.  Generalization capability is particularly important to data efficiency when training data size is limited. For \textbf{model capacity}, we measure the ability to fit large training datasets. When training data is abundant and overfitting is not an issue, the model with higher capacity will achieve better final performance after reasonable training steps. 
Note that, since simply increasing the model size can lead to higher model capacity, to perform a meaningful comparison, we make sure the model sizes of the 5 variants are comparable.

\begin{figure}[!ht]
    \centering
    \begin{subfigure}[b]{0.5\textwidth}
        \centering
        \includegraphics[width=\textwidth]{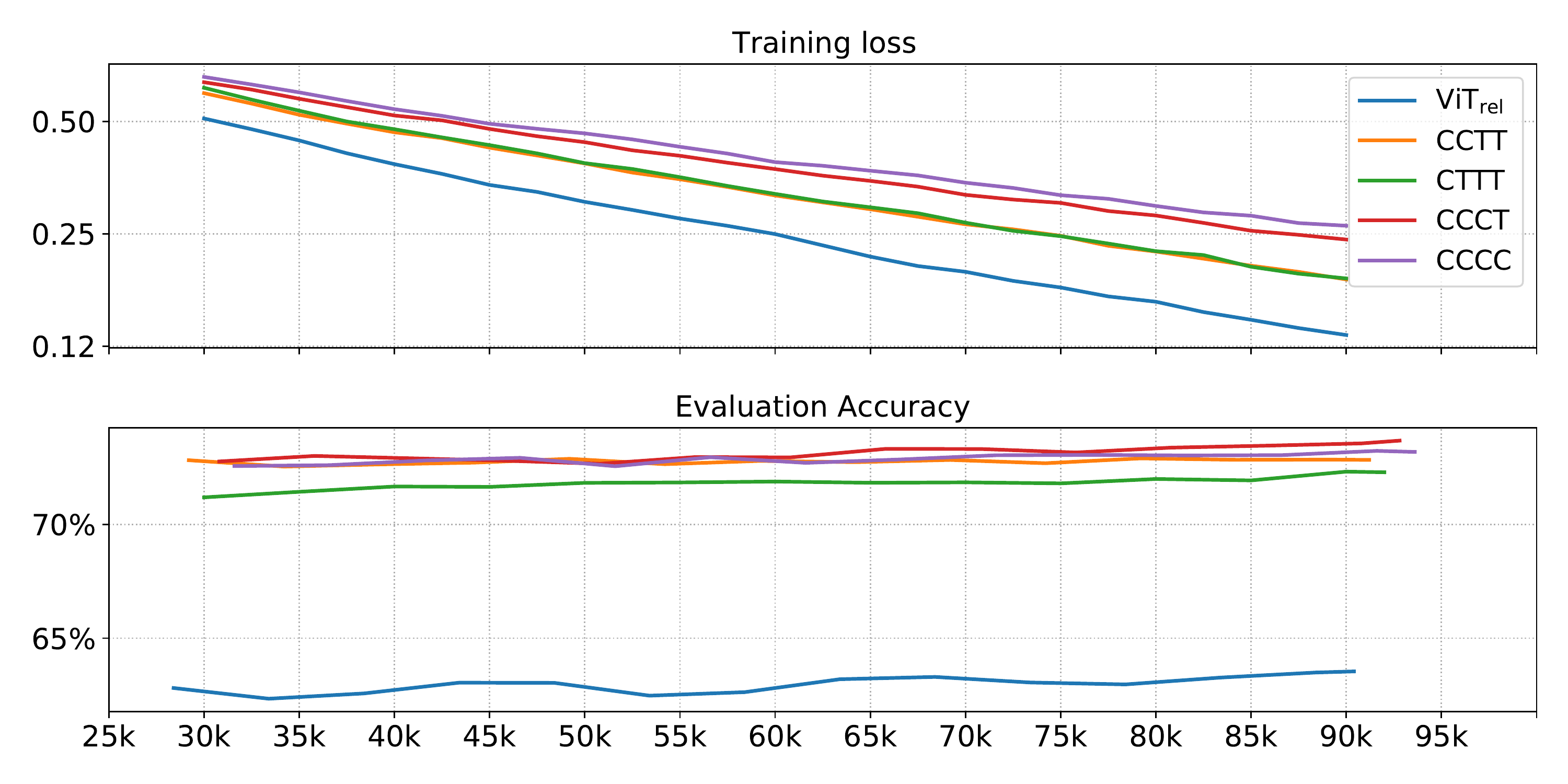}
        \caption{ImageNet-1K}
        \label{fig:i1k}
    \end{subfigure}%
    \begin{subfigure}[b]{0.5\textwidth}
        \centering
        \includegraphics[width=\textwidth]{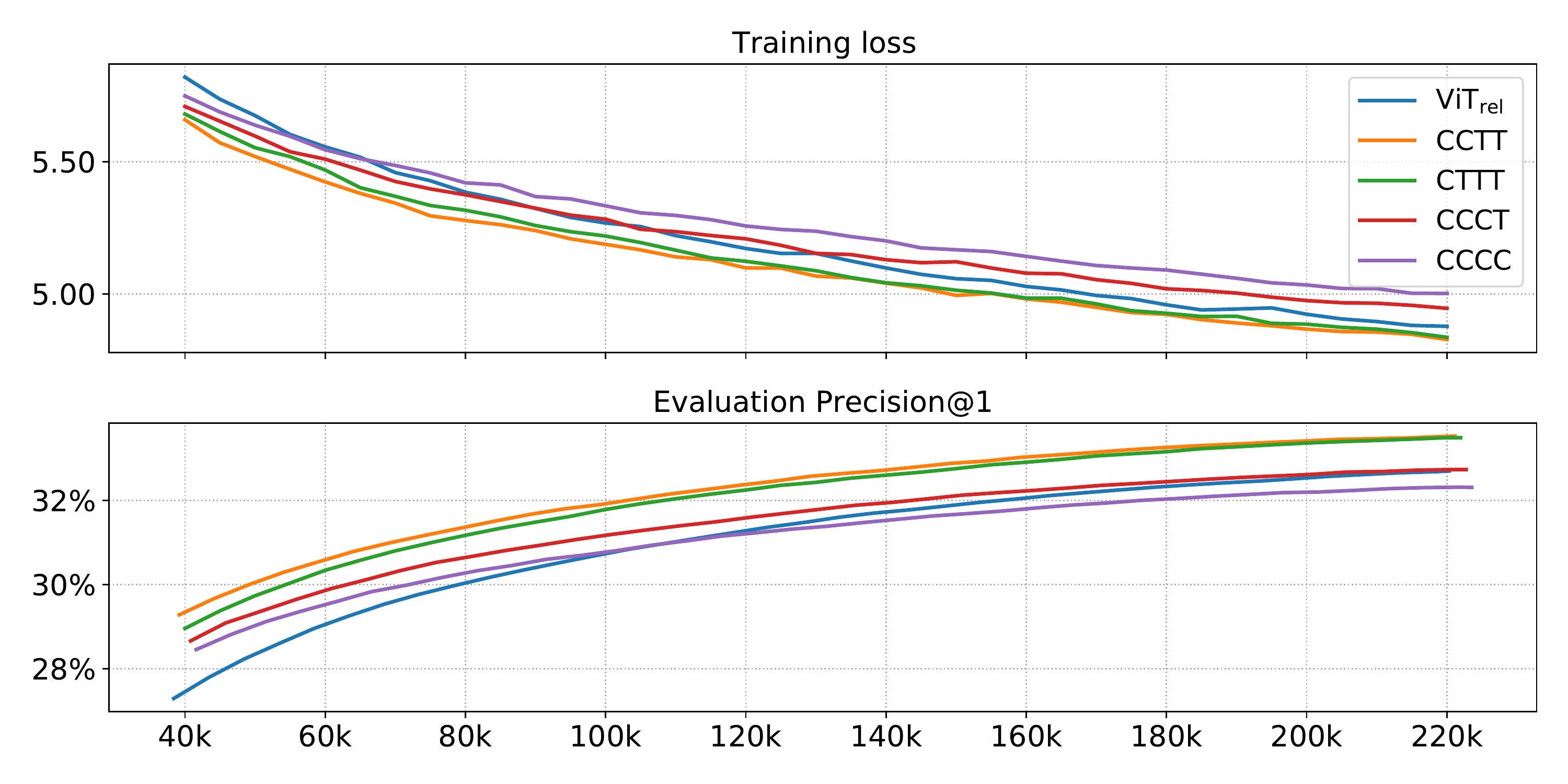}
        \caption{JFT}
        \label{fig:jft}
    \end{subfigure}
    \caption{Comparison for model generalization and capacity under different data size. For fair comparison, all models have similar parameter size and computational cost.}
    \label{fig:verticle_layout}
\end{figure}

To compare the generalization and model capacity, we train different variants of hybrid models on ImageNet-1K (1.3M) and JFT (>300M) dataset for 300 and 3 epochs respectively, both without any regularization or augmentation. The training loss and evaluation accuracy on both datasets are summarized in Figure \ref{fig:verticle_layout}.
\begin{itemize}[leftmargin=*,itemsep=0pt,topsep=0pt,partopsep=0pt]
\item From the ImageNet-1K results, a key observation is that, in terms of \textit{generalization capability} (i.e., gap between train and evaluation metrics), we have
\[
    \textsc{C-C-C-C} \approx \textsc{C-C-C-T} \geq \textsc{C-C-T-T} > \textsc{C-T-T-T} \gg \textsc{ViT$_\text{rel}$}.
\]
Particularly, \textsc{ViT$_\text{rel}$} is significantly worse than variants by a large margin, which we conjecture is related to the lack of proper low-level information processing in its aggressive down-sampling Stem.
Among the multi-stage variants, the overall trend is that the more convolution stages the model has, the smaller the generalization gap is.

\item As for \textit{model capacity}, from the JFT comparison, both the train and evaluation metrics at the end of the training suggest the following ranking:
\[
    \textsc{C-C-T-T} \approx \textsc{C-T-T-T} > \textsc{ViT$_\text{rel}$} > \textsc{C-C-C-T} > \textsc{C-C-C-C}.
\]
Importantly, this suggests that simply having more Transformer blocks does NOT necessarily mean higher capacity for visual processing.
On one hand, while initially worse, \textsc{ViT$_\text{rel}$} ultimately catch up with the two variants with more MBConv stages, indicating the capacity advantage of Transformer blocks.
On the other hand, both $\textsc{C-C-T-T}$ and $\textsc{C-T-T-T}$ clearly outperforming $\textsc{ViT$_\text{rel}$}$ suggest that the ViT stem with an aggressive stride may have lost too much information and hence limit the model capacity.
More interestingly, the fact that $\textsc{C-C-T-T} \approx \textsc{C-T-T-T}$ indicates the for processing low-level information, static local operations like convolution could be as capable as adaptive global attention mechanism, while saving computation and memory usage substantially.
\end{itemize}

Finally, to decide between $\textsc{C-C-T-T}$ and $\textsc{C-T-T-T}$, we conduct another \textbf{transferability} test\footnote{Rigorously speaking, this test examines not only the transferability but also the generalization.} --- we finetune the two JFT pre-trained models above on ImageNet-1K for 30 epochs and compare their transfer performances.
From Table \ref{tab:transfer}, it turns out that $\textsc{C-C-T-T}$ achieves a clearly better transfer accuracy than $\textsc{C-T-T-T}$, despite the same pre-training performance.
\begin{table}[!ht]
    \centering
    \vspace{-1em}
    \caption{Transferability test results.}
    \begin{tabular}{c|cc}
    \toprule
        Metric & \textsc{C-C-T-T} & \textsc{C-T-T-T} \\
        \cmidrule(r){1-1}\cmidrule(l){2-3}  
        Pre-training Precision@1 (JFT) & 34.40 & 34.36 \\
        Transfer Accuracy 224x224  & \bf 82.39 & 81.78 \\
        Transfer Accuracy 384x384  & \bf 84.23 & 84.02 \\
    \bottomrule
    \end{tabular}
    \label{tab:transfer}
\end{table}
\\
Taking generalization, model capacity, transferability and efficiency into consideration, we adapt the \textsc{C-C-T-T} multi-stage layout for \name.
More model details are included in Appendix \ref{sec:appendix_model_detail}.

\section{Related Work}
\label{sec:related}

\paragraph{Convolutional network building blocks.} Convolutional Networks (ConvNets) have been the dominating neural architectures for many computer vision tasks. Traditionally, regular convolutions, such as ResNet blocks~\cite{he2016deep}, are popular in large-scale ConvNets; in contrast, depthwise convolutions~\cite{sepconv14} are popular in mobile platforms due to its lower computational cost and smaller parameter size~\cite{sandler2018mobilenetv2}. Recent works show that an improved inverted residual bottlenecks (MBConv~\cite{sandler2018mobilenetv2,tan2019mnasnet}), which is built upon depthwise convolutions, can achieve both high accuracy and better efficiency~\cite{efficientnet19,tan2021efficientnetv2}.
As discussed in Section \ref{sec:model}, due to the strong connection between MBConv and Transformer blocks , this paper mostly employs MBConv as convolution building blocks.

\paragraph{Self-attention and Transformers.} With the key ingredients of self-attention, Transformers have been widely adopted for neural language processing and speech understanding.
As an early work, stand-alone self-attention network~\cite{ramachandran2019stand} shows self-attention alone can work well for different vision tasks, though with some practical difficulties.
Recently, ViT~\cite{dosovitskiy2020image} applies a vanilla Transformer to ImageNet classification, and achieves impressive results after pre-training on a large-scale JFT dataset. However, ViT still largely lags behind state-of-the-art ConvNets when training data is limited. Following that, many recent works have been focused on improving vision Transformers for data efficiency and model efficiency.
For a more comprehensive review of vision Transformers, we refer readers to the dedicated surveys~\cite{han2020survey,khan2021transformers}.

\paragraph{Relative attention.} Under the general name of relative attention, there have been various variants in literature~\cite{shaw2018self,huang2018music,dai2019transformer,ramachandran2019stand,tsai2019transformer,raffel2019exploring}.
Generally speaking, we can separate them into two categories: (a) the input-dependent version where the extra relative attention score is a function of the input states $f(x_i, x_j, i - j)$, and (b) the input-independent version $f(i - j)$.
The variant in \name belongs to the input-independent version, and is similar to the one used in T5~\cite{raffel2019exploring}, but unlike T5, we neither share the relative attention parameters across layers nor use the bucketing mechanism.
As a benefit of the input independence, obtaining $f(i - j)$ for all $(i, j)$ pairs is computationally much cheaper than the input-dependent version on TPU.
In addition, at inference time, this only needs to be computed once and cached for future use.
A recent work~\cite{liu2021swin} also utilizes such an input-independent parameterization, but it restricts the receptive field to a local window.

\paragraph{Combining convolution and self-attention.}
The idea of combining convolution and self-attention for vision recognition is not new.
A common approach is to augment the ConvNet backbone with explicit self-attention or non-local modules~\cite{wang2018non,bello2019attention,srinivas2021bottleneck,shen2021efficient}, or to replace certain convolution layers with standard self-attention~\cite{srinivas2021bottleneck} or a more flexible mix of linear attention and convolution~\cite{bello2021lambdanetworks}. 
While self-attention usually improves the accuracy, they often come with extra computational cost and hence are often regarded as an add-on to the ConvNets, similar to squeeze-and-excitation~\cite{hu2018squeeze} module.
In comparison, after the success of ViT and ResNet-ViT~\cite{dosovitskiy2020image}, another popular line of research starts with a Transformer backbone and tries to incorporate explicit convolution or some desirable properties of convolution into the Transformer backbone~\cite{yuan2021tokens,graham2021levit,wu2021cvt,liu2021swin,vaswani2021scaling,yuan2021incorporating,wang2021pyramid}.

While our work also belongs to this category, we show that our relative attention instantiation is a natural mixture of depthwise convolution and content-based attention with minimum additional cost.
More importantly, starting from the perspectives of generalization and model capacity, we take a systematic approach to the vertical layout design and show how and why different network stages prefer different types of layers.
Therefore, compared to models that simply use an off-the-shelf ConvNet as the stem layer, such as ResNet-ViT~\cite{dosovitskiy2020image}, \name also scales the Convolution stage (S2) when the overall size increases.
On the other hand, compared to models employing local attention~\cite{liu2021swin,vaswani2021scaling}, \name consistently uses full attention for S3 \& S4 to ensure the model capacity, as S3 occupies the majority of the computation and parameters.

\section{Experiments}
\label{sec:experiments}
In this section, we compare \name with previous results under comparable settings.
For completeness, all the hyper-parameters not mentioned here are included in Appendix \ref{sec:appendix_hyper}.

\subsection{Experiment Setting}

\paragraph{\name model family.} To compare with existing models of different sizes, we also design a family of \name models as summarized in Table \ref{tab:model_family}.
Overall, we always double the number of channels from \texttt{S1} to \texttt{S4}, while ensuring the width of the Stem \texttt{S0} to be smaller or equal to that of \texttt{S1}.
Also, for simplicity, when increasing the depth of the network, we only scale the number of blocks in \texttt{S2} and \texttt{S3}.
\begin{table}[!ht]
\small
    \centering
    \caption{\texttt{L} denotes the number of blocks and \texttt{D} denotes the hidden dimension (\#channels). For all Conv and MBConv blocks, we always use the kernel size 3. For all Transformer blocks, we set the size of each attention head to 32, following \cite{liu2021swin}. The expansion rate for the inverted bottleneck is always 4 and the expansion (shrink) rate for the SE is always 0.25.}
    \begin{tabular}{l|c|l@{\hspace{6pt}}l|l@{\hspace{6pt}}l|l@{\hspace{6pt}}l|l@{\hspace{6pt}}l|l@{\hspace{6pt}}l}
    \toprule
    \bf Stages & \bf Size 
    & \multicolumn{2}{c|}{\bf \name-0} 
    & \multicolumn{2}{c|}{\bf \name-1} 
    & \multicolumn{2}{c|}{\bf \name-2} 
    & \multicolumn{2}{c|}{\bf \name-3} 
    & \multicolumn{2}{c}{\bf \name-4} \\
    \midrule
    \texttt{S0}-Conv & $\sfrac{1}{2}$              
        & \texttt{L=2} & \texttt{D=64}
        & \texttt{L=2} & \texttt{D=64}
        & \texttt{L=2} & \texttt{D=128}
        & \texttt{L=2} & \texttt{D=192} 
        & \texttt{L=2} & \texttt{D=192} \\
    \texttt{S1}-MbConv & $\sfrac{1}{4}$              
        & \texttt{L=2} & \texttt{D=96}
        & \texttt{L=2} & \texttt{D=96}
        & \texttt{L=2} & \texttt{D=128} 
        & \texttt{L=2} & \texttt{D=192} 
        & \texttt{L=2} & \texttt{D=192} \\
    \texttt{S2}-MBConv & $\sfrac{1}{8}$            
        & \texttt{L=3} & \texttt{D=192}
        & \texttt{L=6} & \texttt{D=192}
        & \texttt{L=6} & \texttt{D=256}
        & \texttt{L=6} & \texttt{D=384} 
        & \texttt{L=12} & \texttt{D=384} \\
    \texttt{S3}-TFM$_\text{Rel}$ & $\sfrac{1}{16}$ 
        & \texttt{L=5}  & \texttt{D=384}
        & \texttt{L=14} & \texttt{D=384}
        & \texttt{L=14} & \texttt{D=512}
        & \texttt{L=14} & \texttt{D=768}
        & \texttt{L=28} & \texttt{D=768} \\
    \texttt{S4}-TFM$_\text{Rel}$ & $\sfrac{1}{32}$ 
        & \texttt{L=2} & \texttt{D=768}
        & \texttt{L=2} & \texttt{D=768}
        & \texttt{L=2} & \texttt{D=1024}
        & \texttt{L=2} & \texttt{D=1536}
        & \texttt{L=2} & \texttt{D=1536} \\
    \bottomrule
    \end{tabular}
    \label{tab:model_family}
    \vspace{-1em}
\end{table}
\paragraph{Evaluation Protocol.} Our experiments focus on image classification. 
To evaluate the performance of the model across different data sizes, we utilize three datasets of increasingly larger sizes, namely ImageNet-1K (1.28M images), ImageNet-21K (12.7M images) and JFT (300M images).
Following previous works, we first pre-train our models on each of the three datasets at resolution 224 for 300, 90 and 14 epochs respectively.
Then, we finetune the pre-trained models on ImageNet-1K at the desired resolutions for 30 epochs and obtain the corresponding evaluation accuracy.
One exception is the ImageNet-1K performance at resolution 224, which can be directly obtained at the end of pre-training.
Note that similar to other models utilizing Transformer blocks, directly evaluating models pre-trained on ImageNet-1K at a larger resolution without finetuning usually leads to performance drop.
Hence, finetuning is always employed whenever input resolution changes.

\paragraph{Data Augmentation \& Regularization.} In this work, we only consider two widely used data augmentations, namely RandAugment~\cite{cubuk2020randaugment} and MixUp~\cite{zhang2017mixup}, and three common techniques, including stochastic depth~\cite{huang2016deep}, label smoothing~\cite{szegedy2016rethinking} and weight decay~\cite{loshchilov2017decoupled}, to regularize the model.
Intuitively, the specific hyper-parameters of the augmentation and regularization methods depend on model size and data scale, where strong regularization is usually applied for larger models and smaller dataset.

Under the general principle, a complication under the current paradigm is how to adjust the regularization for pre-training and finetuning as data size can change.
Specifically, we have an interesting observation that if a certain type of augmentation is entirely disabled during pre-training, simply turning it on during fine-tuning would most likely harm the performance rather than improving.
We conjecture this could be related to data distribution shift.
As a result, for certain runs of the proposed model, we deliberately apply RandAugment and stochastic depth of a small degree when pre-training on the two larger datasets, ImageNet21-K and JFT.
Although such regularization can harm the pre-training metrics, this allows more versatile regularization and augmentation during finetuning, leading to improved down-stream performances.

\subsection{Main Results}
\begin{figure}[!h]
\centering
\begin{minipage}{.48\textwidth}
    \centering
    \includegraphics[width=\textwidth]{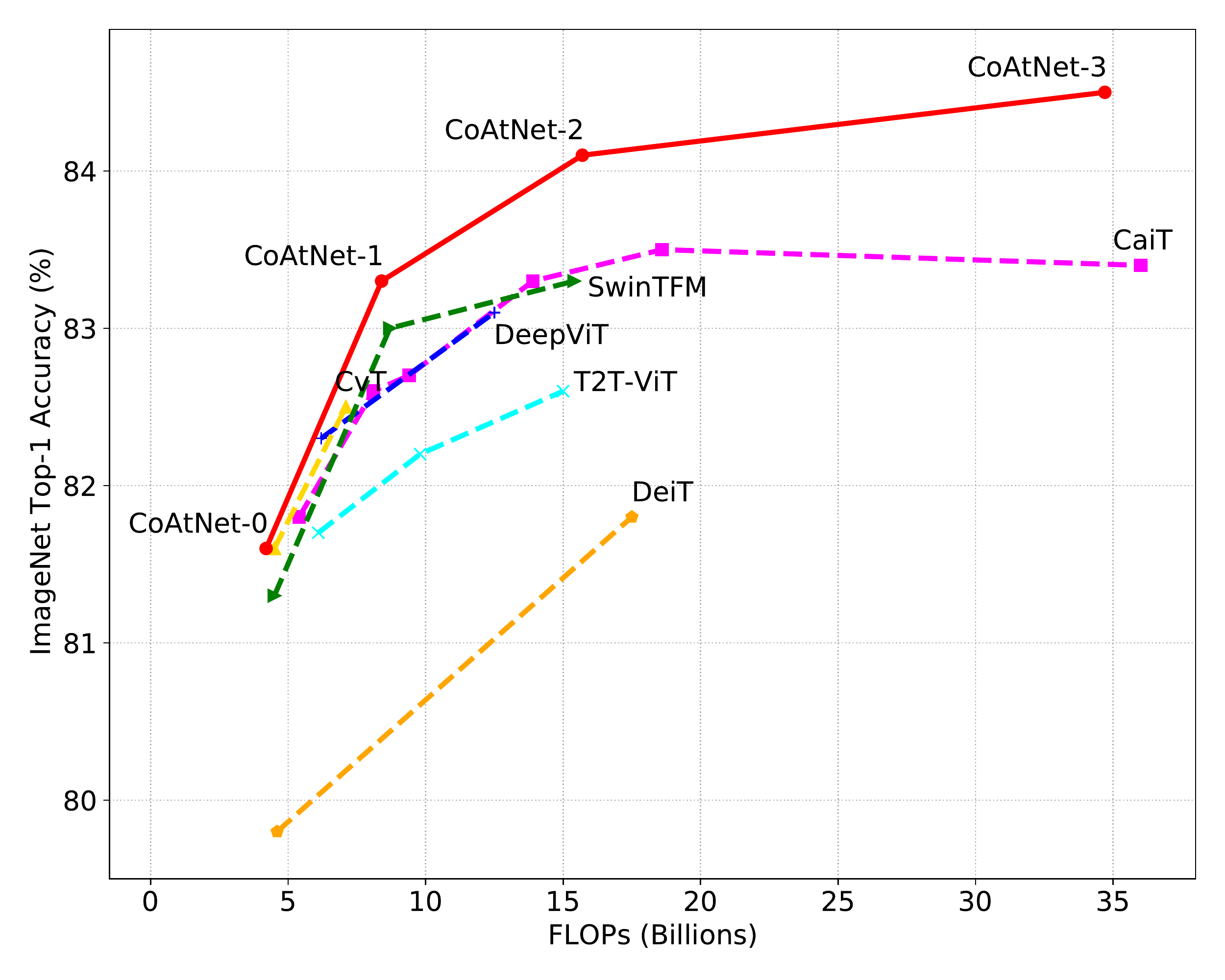}
    \caption{Accuracy-to-FLOPs scaling curve under ImageNet-1K only setting at 224x224.}
    \label{fig:1k_flops}
\end{minipage}\hfill
\begin{minipage}{.48\textwidth}
    \centering
    \includegraphics[width=\textwidth]{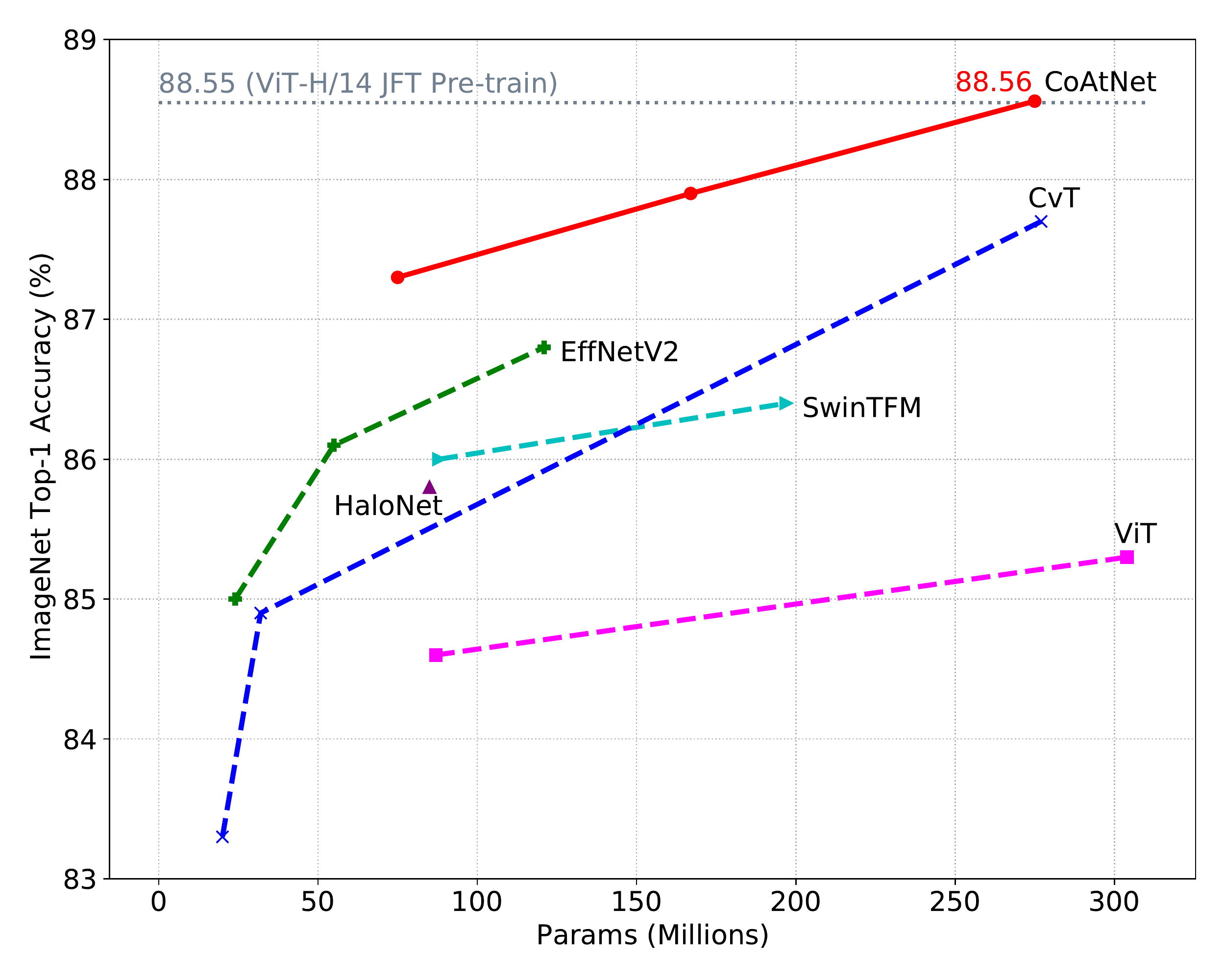}
    \caption{Accuracy-to-Params scaling curve under ImageNet-21K $\Rightarrow$ ImageNet-1K setting.}
    \label{fig:21k_params}
\end{minipage}
\vspace{-1em}
\end{figure}

\begin{table}[!ht]
\small
    \centering
    \vspace{-1em}
    \caption{Model performance on ImageNet.  \texttt{1K only} denotes training on ImageNet-1K only; \texttt{21K+1K} denotes pre-training on ImageNet-21K and finetuning on ImageNet-1K; \texttt{PT-RA} denotes applying RandAugment during 21K pre-training, and \texttt{E150} means 150 epochs of 21K pre-training, which is longer than the standard 90 epochs. More results are in Appendix \ref{sec:appendix_complete_comparison}.}
    \begin{tabular}{c l c c c c c}
        \toprule
        \multicolumn{2}{c}{\bf Models} & \bf Eval Size & \bf \#Params & \bf \#FLOPs & \multicolumn{2}{c}{\bf ImageNet Top-1 Accuracy} \\
        \midrule
        \multicolumn{5}{c}{} & 1K only & 21K+1K \vspace{0.5em} \\
        \multirow{3}{*}{Conv Only}
        & EfficientNet-B7 & $600^2$ & 66M & 37B & 84.7 & -\\
        & EfficientNetV2-L & $480^2$ & 121M & 53B & 85.7 & 86.8\\
        & NFNet-F3 & $416^2$ & 255M & 114.8B & 85.7 & - \\
        & NFNet-F5 & $544^2$ & 377M & 289.8B & \bf 86.0 & - \\
        \cmidrule(r){1-2} \cmidrule(lr){3-3} \cmidrule(lr){4-5} \cmidrule(l){6-6} \cmidrule(l){7-7}
        \multirow{4}{*}{ViT-Stem TFM} 
        & DeiT-B  & $384^2$ & 86M  & 55.4B & 83.1 & - \\
        & ViT-L/16 & $384^2$ & 304M  & 190.7B & -  & 85.3 \\
        & CaiT-S-36 & $384^2$ & 68M & 48.0B & 85.0 & - \\
        & DeepViT-L & $224^2$ & 55M & 12.5B &  83.1 & - \\
        \cmidrule(r){1-2} \cmidrule(lr){3-3} \cmidrule(lr){4-5} \cmidrule(l){6-6} \cmidrule(l){7-7}
        \multirow{3}{*}{Multi-stage TFM} 
        & Swin-B  & $384^2$ & 88M  & 47.0B & 84.2  & 86.0 \\
        & Swin-L  & $384^2$ & 197M & 103.9B & -  & 86.4 \\
        \cmidrule(r){1-2} \cmidrule(lr){3-3} \cmidrule(lr){4-5} \cmidrule(l){6-6} \cmidrule(l){7-7}
        \multirow{5}{*}{Conv+TFM} 
        & BotNet-T7  & $384^2$ & 75.1M & 45.8B & 84.7 & - \\
        & LambdaResNet-420 & $320^2$ & - & - & 84.8 & - \\
        & T2T-ViT-24 & $224^2$ & 64.1M & 15.0B & 82.6 & - \\
        & CvT-21   & $384^2$ & 32M &  24.9B & 83.3 & - \\
        & CvT-W24   & $384^2$ & 277M &  193.2B & - & \bf 87.7 \\
        \midrule
        \multirow{16}{*}{\makecell{\bf Conv+TFM \\ \bf (ours)}} 
        & \name-0 & $224^2$ & 25M  & 4.2B & 81.6 & - \\
        & \name-1 & $224^2$ & 42M  & 8.4B & 83.3 & -  \\
        & \name-2 & $224^2$ & 75M  & 15.7B & 84.1 & 87.1 \\
        & \name-3 & $224^2$ & 168M & 34.7B & 84.5 & 87.6 \\
        \cmidrule(lr){2-2} \cmidrule(lr){3-3} \cmidrule(lr){4-5} \cmidrule(l){6-6} \cmidrule(l){7-7}
        & \name-0 & $384^2$ & 25M  & 13.4B & 83.9  & - \\
        & \name-1 & $384^2$ & 42M  & 27.4B & 85.1  & - \\
        & \name-2 & $384^2$ & 75M  & 49.8B & 85.7 & 87.1 \\
        & \name-3 & $384^2$ & 168M & 107.4B & 85.8 & 87.6\\
        & \name-4 & $384^2$ & 275M & 189.5B & - & 87.9 \\
        & \quad + PT-RA & $384^2$ & 275M & 189.5B & - & 88.3 \\
        & \quad + PT-RA-E150 & $384^2$ & 275M & 189.5B & - & 88.4 \\
        \cmidrule(lr){2-2} \cmidrule(lr){3-3} \cmidrule(lr){4-5} \cmidrule(l){6-6} \cmidrule(l){7-7}
        & \name-2 & $512^2$ & 75M  & 96.7B & 85.9 & 87.3 \\
        & \name-3 & $512^2$ & 168M & 203.1B  & \bf 86.0 & 87.9 \\
        & \name-4 & $512^2$ & 275M & 360.9B & - &  88.1 \\
        & \quad + PT-RA & $512^2$ & 275M & 360.9B & - & 88.4 \\
        & \quad + PT-RA-E150 & $512^2$ & 275M & 360.9B & - & \bf 88.56 \\
        \bottomrule
    \end{tabular}
    \label{tab:i1k21k_result}
\end{table}
\begin{table}[!htbp]
    \centering
    \vspace{-1em}
    \caption{Performance Comparison on large-scale JFT dataset. \texttt{TPUv3-core-days} denotes the pre-training time, \textit{Top-1 Accuracy} denotes the finetuned accuracy on ImageNet. Note that the last 3 rows use a larger dataset JFT-3B~\cite{zhai2021scaling} for pre-training, while others use JFT-300M~\cite{sun2017revisiting}.
    See Appendix \ref{sec:appendix_hyper} for the size details of \name-5/6/7. 
    $^\dagger$: Down-sampling in the MBConv block is achieved by stride-2 Depthwise Convolution.
    $^\diamond$: ViT-G/14 computation consumption is read from Fig. 1 of the paper~\cite{zhai2021scaling}.}
    \begin{tabular}{ccccc|c}
    \midrule 
        \bf Models & \bf Eval Size & \bf \#Params & \bf \#FLOPs & \bf TPUv3-core-days & \bf Top-1 Accuracy \\ 
        \midrule
        ResNet + ViT-L/16 & $384^2$ & 330M & - & - & 87.12 \\
        ViT-L/16  & $512^2$ & 307M & 364B   & 0.68K & 87.76 \\
        ViT-H/14  & $518^2$ & 632M & 1021B  & 2.5K  & 88.55 \\
        NFNet-F4+ & $512^2$ & 527M & 367B   & 1.86K & 89.2 \\
        \midrule
        \name-3$^\dagger$ & $384^2$ & 168M & 114B & 0.58K & 88.52 \\
        \name-3$^\dagger$ & $512^2$ & 168M & 214B & 0.58K & 88.81 \\
        \name-4   & $512^2$ & 275M & 361B   & 0.95K & 89.11 \\
        \name-5   & $512^2$ & 688M & 812B   & 1.82K & 89.77 \\
        \midrule
        ViT-G/14  & $518^2$ & 1.84B & 5160B & >30K$^\diamond$ & 90.45 \\
        \name-6   & $512^2$ & 1.47B & 1521B & 6.6K  & 90.45 \\
        \name-7   & $512^2$ & 2.44B & 2586B & 20.1K & \bf 90.88 \\
        \bottomrule
    \end{tabular}
    \label{tab:jft_result}
    \vspace{-1em}
\end{table}

\paragraph{ImageNet-1K} The experiment results with only the ImageNet-1K dataset are shown in Table \ref{tab:i1k21k_result}.
Under similar conditions, the proposed \name models not only outperform ViT variants, but also match the best convolution-only architectures, i.e., EfficientNet-V2 and NFNets.
Additionally, we also visualize the all results at resolution 224x224 in Fig. \ref{fig:1k_flops}.
As we can see, \name scales much better than previous model with attention modules.

\paragraph{ImageNet-21K} As we can see from Table \ref{tab:i1k21k_result} and Fig. \ref{fig:21k_params}, when ImageNet-21K is used for pre-training, the advantage of \name becomes more obvious, substantially outperforming all previous models.
Notably, the best \name variant achieves a top-1 accuracy of 88.56\%, matching the ViT-H/14 performance of 88.55\%, which requires pre-training the 2.3x larger ViT model on a 23x larger proprietary weakly labeled dataset (JFT) for 2.2x more steps.
This marks a dramatic improvement in both data efficiency and computation efficiency.

\paragraph{JFT} Finally, in Table \ref{tab:jft_result}, we further evaluate \name under the large-scale data regime with JFT-300M and JFT-3B.
Encouragingly, our \name-4 can almost match the best previous performance with JFT-300M set by NFNet-F4+, while being 2x more efficient in terms of both TPU training time and parameter count.
When we scale up the model to consume similar training resource as NFNet-F4+, \name-5 reaches 89.77\% on top-1 accuracy, outperforming previous results under comparable settings.

Moreover, as we further push the training resource towards the level used by ViT-G/14 and utilize the same JFT-3B dataset of an even larger size~\cite{zhai2021scaling}, with over 4x less computation, CoAtNet-6 is able to match the performance of ViT-G/14 of 90.45\%, and with 1.5x less computation, CoAtNet-7 achieves 89.77\% on top-1 accuracy 90.88\%, achieving the new state-of-the-art performance.

\subsection{Ablation Studies}
\label{sec:ablation}
In this section, we will ablate our design choices for  \name.

Firstly, we study the importance of the relative attention from combining convolution and attention into a single computation unit.
Specifically, we compare two models, one with the relative attention and the other without, under both the ImageNet-1K alone and ImageNet-21K transfer setting.
As we can see from Table \ref{tab:rel_attn_ablation}, when only the ImageNet-1K is used, relative attention clearly outperforms the standard attention, indicating a better generalization.
In addition, under the ImageNet-21K transfer setting, the relative attention variant achieves a substantially better transfer accuracy, despite their very close pre-training performances.
This suggests the main advantage of relative attention in visual processing is not in higher capacity but in better generalization.

\begin{table}[!ht]
\centering
\vspace{-1em}
\caption{Ablation on relative attention.}
\begin{tabular}{cc|cc}
    \toprule
    \bf Setting & \bf Metric & \bf With Rel-Attn & \bf Without Rel-Attn  \\
    \midrule
    \multirow{2}{*}{ImageNet-1K} 
    & Accuracy ($224^2$) & 84.1 & 83.8 \\
    & Accuracy ($384^2$) & 85.7 & 85.3 \\
    \midrule
    \multirow{2}{*}{\makecell{ImageNet-21K \\ $\Rightarrow$ ImageNet-1K} } 
    & Pre-train Precision@1 ($224^2$) & 53.0 & 52.8 \\
    & Finetune Accuracy ($384^2$)     & 87.9 & 87.4 \\
    \bottomrule
\end{tabular}
\label{tab:rel_attn_ablation}
\end{table}

\begin{table}[!ht]
    \centering
    \vspace{-1em}
    \caption{Ablation on architecture layout.}
    \begin{tabular}{c l c c}
    \toprule
    \bf Setting & \multicolumn{1}{c}{\bf Models} & \bf Layout & \bf Top-1 Accuracy \\
    \midrule
    \multirow{3}{*}{ImageNet-1K} 
    & V0: \name-2 & \texttt{[2, 2, 6, 14, 2]} & 84.1 \\
    & V1: \texttt{S2} $\Leftarrow$ \texttt{S3} & \texttt{[2, 2, 2, 18, 2]} & 83.4 \\
    & V2: \texttt{S2} $\Rightarrow$ \texttt{S3} & \texttt{[2, 2, 8, 12, 2]} & 84.0 \\
    \midrule
    \multirow{2}{*}{\makecell{ImageNet-21K \\ $\Rightarrow$ ImageNet-1K}}
    & V0: \name-3 & \texttt{[2, 2, 6, 14, 2]} & 53.0 $\to$ 87.6 \\
    & V1: \texttt{S2} $\Leftarrow$ \texttt{S3} & \texttt{[2, 2, 2, 18, 2]} &  53.0 $\to$ 87.4 \\
    \bottomrule
    \end{tabular}
    \label{tab:layout_ablation}
\end{table}
\begin{table}[!ht]
    \centering
    \vspace{-1em}
    \caption{Ablation on head size and normalization type.}
    \begin{tabular}{c l cc}
    \toprule
    \bf Setting & \multicolumn{1}{c}{\bf Models} & \bf Image Size & \bf Top-1 Accuracy \\
    \midrule
    \multirow{3}{*}{ImageNet-1K} 
    & \name-2                     & $224^2$ & 84.1 \\
    & \quad Head size: 32 $\to$ 64 & $224^2$ & 83.9 \\
    & \quad Norm type: BN $\to$ LN & $224^2$ & 84.1 \\
    \midrule
    \multirow{2}{*}{\makecell{ImageNet-21K \\ $\Rightarrow$ ImageNet-1K}}
    & \name-3  & $384^2$ & 87.9 \\
    & \quad Norm type: BN $\to$ LN & $384^2$ & 87.8 \\
    \bottomrule
    \end{tabular}
    \label{tab:misc_ablation}
\end{table}

Secondly, as \texttt{S2} with MBConv blocks and \texttt{S3} with relative Transformer blocks occupy most of the computation of the \name, a question to ask is how to split the computation between \texttt{S2} (MBConv) and \texttt{S3} (Transformer) to achieve a good performance.
In practice, it boils down to deciding the number of blocks to have in each stage, which we will refer to as ``layout'' design.
For this purpose, we compare a few different layouts that we experimented with in Table \ref{tab:layout_ablation}.
\begin{itemize}[leftmargin=*,topsep=0em,itemsep=0em,partopsep=0pt]
\item If we keep the total number of blocks in \texttt{S2} and \texttt{S3} fixed and vary the number in each stage, we observe that V0 is a sweet spot between V1 and V2.
Basically, having more Transformer blocks in \texttt{S3} generally leads to better performance until the number of MBConv blocks in \texttt{S2} is too small to generalize well.

\item To further evaluate whether the sweet spot also holds in the transfer setting, where a higher capacity is often regarded more important, we further compare V0 and V1 under the ImageNet-21K transferring to ImageNet-1K setup.
Interestingly, despite that V1 and V0 have the same performance during ImageNet-21K pre-training, the transfer accuracy of V1 clearly falls behind V0.
Again, this suggests the importance of convolution in achieving good transferability and generalization.
\end{itemize}

Lastly, we study two choices of model details, namely the dimension of each attention (default to 32) head as well as the type of normalization (default to \texttt{BatchNorm}) used in MBConv blocks.
From Table \ref{tab:misc_ablation}, we can see increasing head size from 32 to 64 can slightly hurt performance, though it actually improves the TPU speed by a significant amount.
In practice, this will be a quality-speed trade-off one can make.
On the other hand, \texttt{BatchNorm} and \texttt{LayerNorm} have almost the same performance, while \texttt{BatchNorm} is 10 - 20\% faster on TPU depending on the per-core batch size.

\section{Conclusion}
\label{sec:conclusion}
In this paper, we systematically study the properties of convolutions and Transformers, which leads to a principled way to combine them into a new family of models named \name. Extensive experiments show that \name enjoys both good generalization like ConvNets and superior model capacity like Transformers, achieving state-of-the-art performances under different data sizes and computation budgets.

Note that this paper currently focuses on ImageNet classification for model development. 
However, we believe our approach is applicable to broader applications like object detection and semantic segmentation.
We will leave them for future work.

\clearpage
\bibliographystyle{unsrt}
\bibliography{reference}

\begin{thebibliography}{10}

\bibitem{alexnet12}
Alex Krizhevsky, Ilya Sutskever, and Geoffrey~E Hinton.
\newblock Imagenet classification with deep convolutional neural networks.
\newblock In {\em Advances in Neural Information Processing Systems}, pages
  1097--1105, 2012.

\bibitem{simonyan2014very}
Karen Simonyan and Andrew Zisserman.
\newblock Very deep convolutional networks for large-scale image recognition.
\newblock In {\em ICLR}, 2015.

\bibitem{he2016deep}
Kaiming He, Xiangyu Zhang, Shaoqing Ren, and Jian Sun.
\newblock Deep residual learning for image recognition.
\newblock In {\em CVPR}, 2016.

\bibitem{szegedy2015going}
Christian Szegedy, Wei Liu, Yangqing Jia, Pierre Sermanet, Scott Reed, Dragomir
  Anguelov, Dumitru Erhan, Vincent Vanhoucke, and Andrew Rabinovich.
\newblock Going deeper with convolutions.
\newblock In {\em Proceedings of the IEEE conference on computer vision and
  pattern recognition}, pages 1--9, 2015.

\bibitem{efficientnet19}
Mingxing Tan and Quoc~V. Le.
\newblock Efficientnet: Rethinking model scaling for convolutional neural
  networks.
\newblock {\em ICML}, 2019.

\bibitem{vaswani2017attention}
Ashish Vaswani, Noam Shazeer, Niki Parmar, Jakob Uszkoreit, Llion Jones,
  Aidan~N Gomez, Lukasz Kaiser, and Illia Polosukhin.
\newblock Attention is all you need.
\newblock {\em arXiv preprint arXiv:1706.03762}, 2017.

\bibitem{devlin2018bert}
Jacob Devlin, Ming-Wei Chang, Kenton Lee, and Kristina Toutanova.
\newblock Bert: Pre-training of deep bidirectional transformers for language
  understanding.
\newblock {\em arXiv preprint arXiv:1810.04805}, 2018.

\bibitem{brown2020language}
Tom~B Brown, Benjamin Mann, Nick Ryder, Melanie Subbiah, Jared Kaplan, Prafulla
  Dhariwal, Arvind Neelakantan, Pranav Shyam, Girish Sastry, Amanda Askell,
  et~al.
\newblock Language models are few-shot learners.
\newblock {\em arXiv preprint arXiv:2005.14165}, 2020.

\bibitem{wang2018non}
Xiaolong Wang, Ross Girshick, Abhinav Gupta, and Kaiming He.
\newblock Non-local neural networks.
\newblock In {\em Proceedings of the IEEE conference on computer vision and
  pattern recognition}, pages 7794--7803, 2018.

\bibitem{bello2019attention}
Irwan Bello, Barret Zoph, Ashish Vaswani, Jonathon Shlens, and Quoc~V Le.
\newblock Attention augmented convolutional networks.
\newblock In {\em Proceedings of the IEEE/CVF International Conference on
  Computer Vision}, pages 3286--3295, 2019.

\bibitem{srinivas2021bottleneck}
Aravind Srinivas, Tsung-Yi Lin, Niki Parmar, Jonathon Shlens, Pieter Abbeel,
  and Ashish Vaswani.
\newblock Bottleneck transformers for visual recognition.
\newblock {\em arXiv preprint arXiv:2101.11605}, 2021.

\bibitem{shen2021efficient}
Zhuoran Shen, Mingyuan Zhang, Haiyu Zhao, Shuai Yi, and Hongsheng Li.
\newblock Efficient attention: Attention with linear complexities.
\newblock In {\em Proceedings of the IEEE/CVF Winter Conference on Applications
  of Computer Vision}, pages 3531--3539, 2021.

\bibitem{dosovitskiy2020image}
Alexey Dosovitskiy, Lucas Beyer, Alexander Kolesnikov, Dirk Weissenborn,
  Xiaohua Zhai, Thomas Unterthiner, Mostafa Dehghani, Matthias Minderer, Georg
  Heigold, Sylvain Gelly, et~al.
\newblock An image is worth 16x16 words: Transformers for image recognition at
  scale.
\newblock {\em arXiv preprint arXiv:2010.11929}, 2020.

\bibitem{deng2009imagenet}
Jia Deng, Wei Dong, Richard Socher, Li-Jia Li, Kai Li, and Li~Fei-Fei.
\newblock Imagenet: A large-scale hierarchical image database.
\newblock In {\em 2009 IEEE conference on computer vision and pattern
  recognition}, pages 248--255. Ieee, 2009.

\bibitem{sun2017revisiting}
Chen Sun, Abhinav Shrivastava, Saurabh Singh, and Abhinav Gupta.
\newblock Revisiting unreasonable effectiveness of data in deep learning era.
\newblock In {\em Proceedings of the IEEE international conference on computer
  vision}, pages 843--852, 2017.

\bibitem{touvron2020training}
Hugo Touvron, Matthieu Cord, Matthijs Douze, Francisco Massa, Alexandre
  Sablayrolles, and Herv{\'e} J{\'e}gou.
\newblock Training data-efficient image transformers \& distillation through
  attention.
\newblock {\em arXiv preprint arXiv:2012.12877}, 2020.

\bibitem{touvron2021going}
Hugo Touvron, Matthieu Cord, Alexandre Sablayrolles, Gabriel Synnaeve, and
  Herv{\'e} J{\'e}gou.
\newblock Going deeper with image transformers.
\newblock {\em arXiv preprint arXiv:2103.17239}, 2021.

\bibitem{zhou2021deepvit}
Daquan Zhou, Bingyi Kang, Xiaojie Jin, Linjie Yang, Xiaochen Lian, Qibin Hou,
  and Jiashi Feng.
\newblock Deepvit: Towards deeper vision transformer.
\newblock {\em arXiv preprint arXiv:2103.11886}, 2021.

\bibitem{tan2021efficientnetv2}
Mingxing Tan and Quoc~V Le.
\newblock Efficientnetv2: Smaller models and faster training.
\newblock {\em ICML}, 2021.

\bibitem{brock2021high}
Andrew Brock, Soham De, Samuel~L Smith, and Karen Simonyan.
\newblock High-performance large-scale image recognition without normalization.
\newblock {\em arXiv preprint arXiv:2102.06171}, 2021.

\bibitem{vaswani2021scaling}
Ashish Vaswani, Prajit Ramachandran, Aravind Srinivas, Niki Parmar, Blake
  Hechtman, and Jonathon Shlens.
\newblock Scaling local self-attention for parameter efficient visual
  backbones.
\newblock {\em arXiv preprint arXiv:2103.12731}, 2021.

\bibitem{liu2021swin}
Ze~Liu, Yutong Lin, Yue Cao, Han Hu, Yixuan Wei, Zheng Zhang, Stephen Lin, and
  Baining Guo.
\newblock Swin transformer: Hierarchical vision transformer using shifted
  windows.
\newblock {\em arXiv preprint arXiv:2103.14030}, 2021.

\bibitem{wu2021cvt}
Haiping Wu, Bin Xiao, Noel Codella, Mengchen Liu, Xiyang Dai, Lu~Yuan, and Lei
  Zhang.
\newblock Cvt: Introducing convolutions to vision transformers.
\newblock {\em arXiv preprint arXiv:2103.15808}, 2021.

\bibitem{graham2021levit}
Ben Graham, Alaaeldin El-Nouby, Hugo Touvron, Pierre Stock, Armand Joulin,
  Herv{\'e} J{\'e}gou, and Matthijs Douze.
\newblock Levit: a vision transformer in convnet's clothing for faster
  inference.
\newblock {\em arXiv preprint arXiv:2104.01136}, 2021.

\bibitem{yuan2021tokens}
Li~Yuan, Yunpeng Chen, Tao Wang, Weihao Yu, Yujun Shi, Francis~EH Tay, Jiashi
  Feng, and Shuicheng Yan.
\newblock Tokens-to-token vit: Training vision transformers from scratch on
  imagenet.
\newblock {\em arXiv preprint arXiv:2101.11986}, 2021.

\bibitem{zhai2021scaling}
Xiaohua Zhai, Alexander Kolesnikov, Neil Houlsby, and Lucas Beyer.
\newblock Scaling vision transformers.
\newblock {\em arXiv preprint arXiv:2106.04560}, 2021.

\bibitem{sandler2018mobilenetv2}
Mark Sandler, Andrew Howard, Menglong Zhu, Andrey Zhmoginov, and Liang-Chieh
  Chen.
\newblock Mobilenetv2: Inverted residuals and linear bottlenecks.
\newblock In {\em Proceedings of the IEEE conference on computer vision and
  pattern recognition}, pages 4510--4520, 2018.

\bibitem{sepconv14}
Laurent Sifre.
\newblock Rigid-motion scattering for image classification.
\newblock {\em Ph.D. thesis section 6.2}, 2014.

\bibitem{mohamed2020data}
Mirgahney Mohamed, Gabriele Cesa, Taco~S Cohen, and Max Welling.
\newblock A data and compute efficient design for limited-resources deep
  learning.
\newblock {\em arXiv preprint arXiv:2004.09691}, 2020.

\bibitem{shaw2018self}
Peter Shaw, Jakob Uszkoreit, and Ashish Vaswani.
\newblock Self-attention with relative position representations.
\newblock {\em arXiv preprint arXiv:1803.02155}, 2018.

\bibitem{raffel2019exploring}
Colin Raffel, Noam Shazeer, Adam Roberts, Katherine Lee, Sharan Narang, Michael
  Matena, Yanqi Zhou, Wei Li, and Peter~J Liu.
\newblock Exploring the limits of transfer learning with a unified text-to-text
  transformer.
\newblock {\em arXiv preprint arXiv:1910.10683}, 2019.

\bibitem{katharopoulos2020transformers}
Angelos Katharopoulos, Apoorv Vyas, Nikolaos Pappas, and Fran{\c{c}}ois
  Fleuret.
\newblock Transformers are rnns: Fast autoregressive transformers with linear
  attention.
\newblock In {\em International Conference on Machine Learning}, pages
  5156--5165. PMLR, 2020.

\bibitem{choromanski2020rethinking}
Krzysztof Choromanski, Valerii Likhosherstov, David Dohan, Xingyou Song,
  Andreea Gane, Tamas Sarlos, Peter Hawkins, Jared Davis, Afroz Mohiuddin,
  Lukasz Kaiser, et~al.
\newblock Rethinking attention with performers.
\newblock {\em arXiv preprint arXiv:2009.14794}, 2020.

\bibitem{ramachandran2019stand}
Prajit Ramachandran, Niki Parmar, Ashish Vaswani, Irwan Bello, Anselm Levskaya,
  and Jonathon Shlens.
\newblock Stand-alone self-attention in vision models.
\newblock {\em arXiv preprint arXiv:1906.05909}, 2019.

\bibitem{tan2019mnasnet}
Mingxing Tan, Bo~Chen, Ruoming Pang, Vijay Vasudevan, Mark Sandler, Andrew
  Howard, and Quoc~V Le.
\newblock Mnasnet: Platform-aware neural architecture search for mobile.
\newblock In {\em Proceedings of the IEEE/CVF Conference on Computer Vision and
  Pattern Recognition}, pages 2820--2828, 2019.

\bibitem{han2020survey}
Kai Han, Yunhe Wang, Hanting Chen, Xinghao Chen, Jianyuan Guo, Zhenhua Liu,
  Yehui Tang, An~Xiao, Chunjing Xu, Yixing Xu, et~al.
\newblock A survey on visual transformer.
\newblock {\em arXiv preprint arXiv:2012.12556}, 2020.

\bibitem{khan2021transformers}
Salman Khan, Muzammal Naseer, Munawar Hayat, Syed~Waqas Zamir, Fahad~Shahbaz
  Khan, and Mubarak Shah.
\newblock Transformers in vision: A survey.
\newblock {\em arXiv preprint arXiv:2101.01169}, 2021.

\bibitem{huang2018music}
Cheng-Zhi~Anna Huang, Ashish Vaswani, Jakob Uszkoreit, Noam Shazeer, Ian Simon,
  Curtis Hawthorne, Andrew~M Dai, Matthew~D Hoffman, Monica Dinculescu, and
  Douglas Eck.
\newblock Music transformer.
\newblock {\em arXiv preprint arXiv:1809.04281}, 2018.

\bibitem{dai2019transformer}
Zihang Dai, Zhilin Yang, Yiming Yang, Jaime Carbonell, Quoc~V Le, and Ruslan
  Salakhutdinov.
\newblock Transformer-xl: Attentive language models beyond a fixed-length
  context.
\newblock {\em arXiv preprint arXiv:1901.02860}, 2019.

\bibitem{tsai2019transformer}
Yao-Hung~Hubert Tsai, Shaojie Bai, Makoto Yamada, Louis-Philippe Morency, and
  Ruslan Salakhutdinov.
\newblock Transformer dissection: A unified understanding of transformer's
  attention via the lens of kernel.
\newblock {\em arXiv preprint arXiv:1908.11775}, 2019.

\bibitem{bello2021lambdanetworks}
Irwan Bello.
\newblock Lambdanetworks: Modeling long-range interactions without attention.
\newblock {\em arXiv preprint arXiv:2102.08602}, 2021.

\bibitem{hu2018squeeze}
Jie Hu, Li~Shen, and Gang Sun.
\newblock Squeeze-and-excitation networks.
\newblock In {\em Proceedings of the IEEE conference on computer vision and
  pattern recognition}, pages 7132--7141, 2018.

\bibitem{yuan2021incorporating}
Kun Yuan, Shaopeng Guo, Ziwei Liu, Aojun Zhou, Fengwei Yu, and Wei Wu.
\newblock Incorporating convolution designs into visual transformers.
\newblock {\em arXiv preprint arXiv:2103.11816}, 2021.

\bibitem{wang2021pyramid}
Wenhai Wang, Enze Xie, Xiang Li, Deng-Ping Fan, Kaitao Song, Ding Liang, Tong
  Lu, Ping Luo, and Ling Shao.
\newblock Pyramid vision transformer: A versatile backbone for dense prediction
  without convolutions.
\newblock {\em arXiv preprint arXiv:2102.12122}, 2021.

\bibitem{cubuk2020randaugment}
Ekin~D Cubuk, Barret Zoph, Jonathon Shlens, and Quoc~V Le.
\newblock Randaugment: Practical automated data augmentation with a reduced
  search space.
\newblock In {\em Proceedings of the IEEE/CVF Conference on Computer Vision and
  Pattern Recognition Workshops}, pages 702--703, 2020.

\bibitem{zhang2017mixup}
Hongyi Zhang, Moustapha Cisse, Yann~N Dauphin, and David Lopez-Paz.
\newblock mixup: Beyond empirical risk minimization.
\newblock {\em arXiv preprint arXiv:1710.09412}, 2017.

\bibitem{huang2016deep}
Gao Huang, Yu~Sun, Zhuang Liu, Daniel Sedra, and Kilian~Q Weinberger.
\newblock Deep networks with stochastic depth.
\newblock In {\em European conference on computer vision}, pages 646--661.
  Springer, 2016.

\bibitem{szegedy2016rethinking}
Christian Szegedy, Vincent Vanhoucke, Sergey Ioffe, Jon Shlens, and Zbigniew
  Wojna.
\newblock Rethinking the inception architecture for computer vision.
\newblock In {\em Proceedings of the IEEE conference on computer vision and
  pattern recognition}, pages 2818--2826, 2016.

\bibitem{loshchilov2017decoupled}
Ilya Loshchilov and Frank Hutter.
\newblock Decoupled weight decay regularization.
\newblock {\em arXiv preprint arXiv:1711.05101}, 2017.

\bibitem{he2016identity}
Kaiming He, Xiangyu Zhang, Shaoqing Ren, and Jian Sun.
\newblock Identity mappings in deep residual networks.
\newblock In {\em European conference on computer vision}, pages 630--645.
  Springer, 2016.

\bibitem{hendrycks2016gaussian}
Dan Hendrycks and Kevin Gimpel.
\newblock Gaussian error linear units (gelus).
\newblock {\em arXiv preprint arXiv:1606.08415}, 2016.

\bibitem{dai2020funnel}
Zihang Dai, Guokun Lai, Yiming Yang, and Quoc~V Le.
\newblock Funnel-transformer: Filtering out sequential redundancy for efficient
  language processing.
\newblock {\em arXiv preprint arXiv:2006.03236}, 2020.

\end{thebibliography}
\clearpage

\appendix

\section{Appendix}
\label{sec:appendix}

\subsection{Model Details}
\label{sec:appendix_model_detail}
\begin{figure}[!ht]
    \centering
    \includegraphics[width=\textwidth]{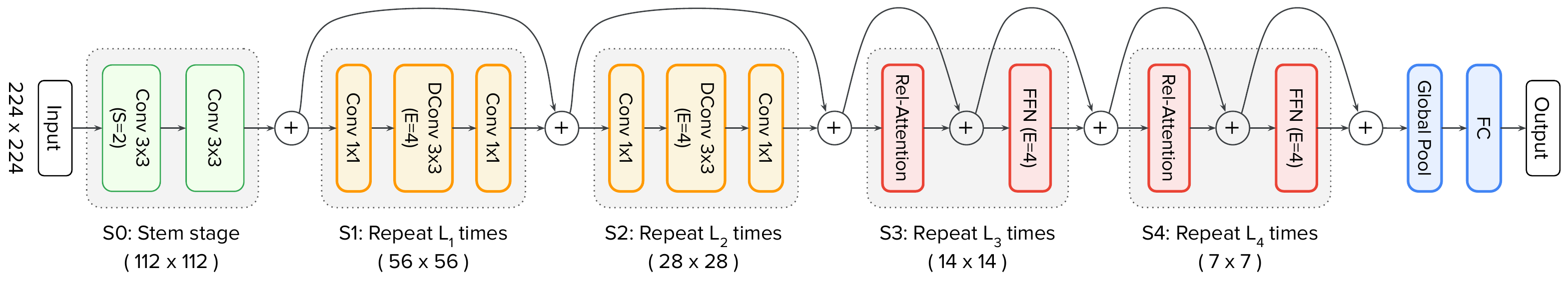}
    \caption{Overview of the proposed \name.}
    \label{fig:overview}
\end{figure}

First of all, the overview of \name is illustrated in Fig. \ref{fig:overview}.
\paragraph{2D Relative Attention} To implement the pre-norm version of relative attention in Eqn. \ref{eqn:rel_attn} for 2D images of size $[H \times W]$, for \textit{each head}, we create a trainable parameter $\rmP$ of size $[(2H - 1) \times (2W - 1)]$, as the maximum distance is $2H - 1$ and $2W - 1$ respectively.
Then, for two spatial locations $(i, j)$ and $(i', j')$, the corresponding relative bias is $P_{i-i'+H,j-j'+W}$ under 1-based indexing.
For implementation, we need to index $H^2W^2$ elements from the $[(2H - 1) \times (2W - 1)]$ matrix.
On TPU, we utilize two einsums, along the height and width axis respectively, to index the relative bias with complexity $O(HW(H + W))$, which is strictly subsumed by the $O(H^2W^2D)$ attention complexity.
On GPUs, the indexing can be done more efficiently with \texttt{gather}, which only requires memory access.
Note that, at inference time, indexing the $H^2W^2$ elements from the $[(2H - 1) \times (2W - 1)]$ matrix can be pre-computed and cached to further increase the throughput.

When finetuned on a larger resolution, we simply use bi-linear interpolation to increase the size $[(2H - 1) \times (2W - 1)]$ to the desired size $[(2H' - 1) \times (2W' - 1)]$ for any $H' > H$ and $W' > W$.

\paragraph{Pre-Activation} To promote homogeneity in the model architecture, we consistently use pre-activation structure~\cite{he2016identity} for both the MBConv and the Transformer block, i.e.,
\[
    \rvx \leftarrow \rvx + \texttt{Module}\left( \texttt{Norm}(\rvx) \right),
\]
where \texttt{Module} denotes the MBConv, Self-Attention or FFN module, while \texttt{Norm} corresponds to \texttt{BatchNorm} for MBConv and \texttt{LayerNorm} for Self-Attention and FFN.
We have experimented with using \texttt{LayerNorm} in the MBConv block, which achieves the same performance while being significantly slower on our accelerator (TPU).
In general, we recommend whichever is faster on your device.
Following the same spirit, Gaussian Error Linear Units (GELUs)~\cite{hendrycks2016gaussian} is used as the activation function in both the MBConv blocks and Transformer blocks.

\paragraph{Down-Sampling} For the first block inside each stage from \texttt{S1} to \texttt{S4}, down-sampling is performed independently for the residual branch and the identity branch.
Specifically, for the Transformer block, the standard max pooling of stride 2 is directly applied to the input states of both branches of the self-attention module, similar to Funnel Transformer~\cite{dai2020funnel}.
Also, a channel projection is applied to the identity branch to enlarge the hidden size.
Hence, the down-sampling self-attention module can be expressed as
\begin{equation}
\label{eqn:attn_pool}
    \rvx \leftarrow \texttt{Proj}(\texttt{Pool}(\rvx)) + \texttt{Attention}\left( \texttt{Pool}(\texttt{Norm}(\rvx)) \right).
\end{equation}
As for the MBConv block, the down-sampling in the residual branch is instead achieved 
by using a stride-2 convolution to the normalized inputs, i.e.,
\begin{equation}
\label{eqn:mbconv_pool}
    \rvx \leftarrow \texttt{Proj}(\texttt{Pool}(\rvx)) + \texttt{Conv}\left(\texttt{DepthConv}\left(\texttt{Conv}\left( \texttt{Norm}(\rvx), \texttt{stride}=2) \right) \right) \right).
\end{equation}
This is different from the standard MBConv where the down-sampling is done by applying stride-2 depthwise convolution to the inverted bottleneck hidden states.
We later found using stride-2 depthwise convolution is helpful but slower when model is small but not so much when model scales, as shown in Table \ref{tab:correction}.
Hence, if not mentioned otherwise, numbers reported in the main text uses the down-sampling implementation in Eqn. \eqref{eqn:mbconv_pool}.
In practice, this could be yet another quality-speed trade-off one can tweak for smaller models.
\begin{table}[!ht]
\small
    \centering
    \caption{The effect of performing down-sampling in first Conv v.s. the Depthwise Conv.}
    \begin{tabular}{l c c c c}
        \toprule
        \bf Models & \bf Eval Size & \bf \#Params & \bf \#FLOPs & \bf ImageNet Top-1 Accuracy \\
        \midrule
        \name-0             & $224^2$ & 25M & 4.2B & 81.6 \\
        \quad Strided DConv & $224^2$ & 25M & 4.6B & 82.0 \\
        \name-1             & $224^2$ & 42M & 8.4B & 83.3 \\
        \quad Strided DConv & $224^2$ & 42M & 8.8B & 83.5 \\
        \name-2             & $224^2$ & 75M & 15.7B & 84.1 \\
        \quad Strided DConv & $224^2$ & 75M & 16.6B & 84.1  \\
        \bottomrule
    \end{tabular}
    \label{tab:correction}
\end{table}

\paragraph{Classification head} Instead of adding an additional \texttt{<cls>} token as in ViT to perform classification, we apply global average pooling to the last-stage output to get the representation for simplicity.

\subsection{Hyper-Parameters}
\label{sec:appendix_hyper}

\begin{table}[!ht]
\small
\caption{Hyper-parameters used in the main experiments. The slash sign `` / '' is used to separate the different hyper-parameters used for various \name model sizes. $^\diamond$: For finetuning the slightly larger \name-3, RandAugment of 2, 20 is used. $^\dagger$: RandAugment of 2, 5 is applied to the PT-RA variants of \name-4 in Table \ref{tab:21k_result}. }
\centering
\begin{tabular}{@{}l@{\hspace{5pt}}|c@{\hspace{5pt}}c|c@{\hspace{5pt}}c|c@{\hspace{5pt}}c@{}}
    \toprule
    \multirow{3}{*}{\bf Hyper-parameter} 
    & \multicolumn{2}{c|}{\bf ImageNet-1K} 
    & \multicolumn{2}{c|}{\bf ImageNet-21K} 
    & \multicolumn{2}{c}{\bf JFT} \\
    & Pre-Training & Finetuning & Pre-Training & Finetuning & Pre-Training & Finetuning \\
    & \multicolumn{2}{c|}{(\name-0/1/2/3)} 
    & \multicolumn{2}{c|}{(\name-2/3/4)} 
    & \multicolumn{2}{c}{(\name-3/4/5)} \\
    \cmidrule(r){1-1} \cmidrule(l){2-3}  \cmidrule(l){4-5}  \cmidrule(l){6-7} 
    Stochastic depth rate 
    & \multicolumn{2}{c|}{0.2 / 0.3 / 0.5 / 0.7} 
    & \multicolumn{2}{c|}{0.3 / 0.5 / 0.7} 
    & 0.0 / 0.1 / 0.0 & 0.1 / 0.3 / 0.2 \\
    Center crop           & True      & False   & True     & False   & True        & False \\
    RandAugment           & 2, 15     & 2, 15$^\diamond$   & \multicolumn{2}{c|}{None / None / 2, 5$^\dagger$} & 2, 5 & 2, 5 \\
    Mixup alpha           & 0.8       & 0.8     & None     & None    & None        & None \\
    Loss type             & Softmax   & Softmax & Sigmoid  & Softmax & Sigmoid     & Softmax \\
    Label smoothing       & 0.1       & 0.1     & 0.0001   & 0.1     & 0.0001      & 0.1 \\
    Train epochs          & 300       & 30      & 90       & 30      & 14          & 30 \\
    Train batch size      & 4096      & 512     & 4096     & 1024    & 4096        & 512 \\
    Optimizer type        & AdamW     & AdamW   & AdamW    & AdamW   & AdamW       & AdamW \\
    Peak learning rate    & 1e-3      & 5e-5    & 1e-3     & 5e-5    & 1e-3 / 5e-4 / 5e-4 & 5e-5 \\
    Min learning rate     & 1e-5      & 5e-5    & 1e-5     & 5e-5    & 1e-5        & 5e-5\\
    Warm-up               & 10K steps & None    & 5 epochs & None    & 20K steps   & None    \\
    LR decay schedule     & Cosine    & None    & Linear   & None    & Linear      & None\\
    Weight decay rate     & 0.05      & 1e-8    & 0.01     & 1e-8    & 0.01        & 1e-8 \\
    Gradient clip         & 1.0       & 1.0     & 1.0      & 1.0     & 1.0         & 1.0 \\
    EMA decay rate        & None      & 0.9999  & None     & 0.9999  & None        & 0.9999 \\
    \bottomrule
\end{tabular}
\label{tab:main_hparam}
\end{table}
The hyper-parameters used for the main experiments presented in Section \ref{sec:experiments} are summarized in Table \ref{tab:main_hparam}.

The model size of \name-5 used in the JFT experiment is summarized in Table \ref{tab:jft_model}.
Different from the standard \name models in Table \ref{tab:model_family}, we set the size of each attention head to 64 rather than 32 for \name-5, as this achieves a better speed-performance trade-off as discussed in Section \ref{sec:ablation}.
\begin{table}[!ht]
    \centering
    \caption{\name-5 model sizes.}
    \begin{tabular}{l|c|ll}
    \toprule
    \bf Stages & \bf Size 
    & \multicolumn{2}{c}{\bf \name-5} \\
    \midrule
    \texttt{S0}-Conv & $\sfrac{1}{2}$              
        & \texttt{L=2} & \texttt{D=192} \\
    \texttt{S1}-MbConv & $\sfrac{1}{4}$              
        & \texttt{L=2} & \texttt{D=256} \\
    \texttt{S2}-MBConv & $\sfrac{1}{8}$            
        & \texttt{L=12} & \texttt{D=512} \\
    \texttt{S3}-TFM$_\text{Rel}$ & $\sfrac{1}{16}$ 
        & \texttt{L=28}  & \texttt{D=1280} \\
    \texttt{S4}-TFM$_\text{Rel}$ & $\sfrac{1}{32}$ 
        & \texttt{L=2} & \texttt{D=2048} \\
    \bottomrule
    \end{tabular}
    \label{tab:jft_model}
\end{table}

For CoAtNet-6 and CoAtNet-7, to reduce the memory consumption, we move $\sfrac{2}{3}$ of the MBConv blocks of S2 into S3 and double its hidden dimension.
While this modification does not change the complexity in terms of FLOPs, this will reduce the activation related memory usage of these MBConv blocks by half, which enables us to build a larger model.
With this adjustment, the S3 becomes a stage of mixed block types and hidden dimensions. 
In addition, we increase the attention head size to 128 further to boost the speed-performance trade-off.
The specific sizes are summarized in Table~\ref{tab:scaled_jft_model}.
Basically, CoAtNet-6 and CoAtNet-7 share the same depth but differ in width.

\begin{table}[!ht]
    \centering
    \caption{Model sizes for the scaled models.}
    \begin{tabular}{l c ll ll}
    \toprule
    \bf Stages & \bf Size 
    & \multicolumn{2}{c}{\bf \name-6} & \multicolumn{2}{c}{\bf \name-7} \\
    \cmidrule(r){1-2} \cmidrule(lr){3-4} \cmidrule(l){5-6}
    \texttt{S0}-Conv & $\sfrac{1}{2}$              
        & \texttt{L=2} & \texttt{D=192} 
        & \texttt{L=2} & \texttt{D=192} \\
    \cmidrule(r){1-2} \cmidrule(lr){3-4} \cmidrule(l){5-6}
    \texttt{S1}-MbConv & $\sfrac{1}{4}$              
        & \texttt{L=2} & \texttt{D=192} 
        & \texttt{L=2} & \texttt{D=256} \\
    \cmidrule(r){1-2} \cmidrule(lr){3-4} \cmidrule(l){5-6}
    \texttt{S2}-MBConv & $\sfrac{1}{8}$            
        & \texttt{L=4} & \texttt{D=384} 
        & \texttt{L=4} & \texttt{D=512} \\
    \cmidrule(r){1-2} \cmidrule(lr){3-4} \cmidrule(l){5-6}
    \texttt{S3}-MBConv & \multirow{2}{*}{$\sfrac{1}{16}$}
        & \texttt{L=8}  & \texttt{D=768} 
        & \texttt{L=8}  & \texttt{D=1024} \\
    \texttt{S3}-TFM$_\text{Rel}$ & 
        & \texttt{L=42}  & \texttt{D=1536} 
        & \texttt{L=42}  & \texttt{D=2048} \\
    \cmidrule(r){1-2} \cmidrule(lr){3-4} \cmidrule(l){5-6}
    \texttt{S4}-TFM$_\text{Rel}$ & $\sfrac{1}{32}$ 
        & \texttt{L=2} & \texttt{D=2048} 
        & \texttt{L=2}  & \texttt{D=3072} \\
    \bottomrule
    \end{tabular}
    \label{tab:scaled_jft_model}
\end{table}

\subsection{Complete Comparison}
\label{sec:appendix_complete_comparison}
\begin{table}[!ht]
\small
    \centering
    \caption{Complete comparison under the ImageNet-1K only setting.}
    \begin{tabular}{c l c c c c}
        \toprule
        \multicolumn{2}{c}{\bf Models} & \bf Eval Size & \bf \#Params & \bf \#FLOPs & \bf Top-1 Accuracy \\
        \midrule
        \multirow{11}{*}{Conv Only}
        & ResNet-RS-152 & $256^2$ & 87M & 31B & 83.0 \\
        & ResNet-RS-420 & $320^2$ & 192M & 128B & 84.4 \\
        \cmidrule(lr){2-2} \cmidrule(lr){3-3} \cmidrule(lr){4-5} \cmidrule(l){6-6}
        & NFNet-F0 & $256^2$ & 72M  & 12.4B & 83.6 \\
        & NFNet-F1 & $320^2$ & 133M & 35.5B & 84.7 \\
        & NFNet-F2 & $352^2$ & 194M & 62.6B & 85.1 \\
        & NFNet-F3 & $416^2$ & 255M & 114.8B & 85.7 \\
        & NFNet-F4 & $512^2$ & 316M & 215.2B & 85.9 \\
        & NFNet-F5 & $544^2$ & 377M & 289.8B & \bf 86.0 \\
        \cmidrule(lr){2-2} \cmidrule(lr){3-3} \cmidrule(lr){4-5} \cmidrule(l){6-6}
        & ENetV2-S & $384^2$ & 24M  & 8.8B & 83.9 \\
        & ENetV2-M & $480^2$ & 55M  & 24B & 85.1 \\
        & ENetV2-L & $480^2$ & 121M & 53B & 85.7 \\
        \cmidrule(r){1-2} \cmidrule(lr){3-3} \cmidrule(lr){4-5} \cmidrule(l){6-6}
        \multirow{11}{*}{ViT-Stem TFM Only} 
        & DeiT-S  & $224^2$ & 22M  & 4.6B & 79.8 \\
        & DeiT-B  & $224^2$ & 86M  & 17.5B & 81.8 \\
        & DeiT-B  & $384^2$ & 86M  & 55.4B & 83.1 \\
        \cmidrule(lr){2-2} \cmidrule(lr){3-3} \cmidrule(lr){4-5} \cmidrule(l){6-6}
        & CaiT-S-24 & $224^2$ & 46.9M & 9.4B & 82.7 \\
        & CaiT-S-36 & $224^2$ & 68.2M & 13.9B & 83.3 \\
        & CaiT-M-24 & $224^2$ & 185.9M & 36.0B & 83.4 \\
        & CaiT-S-24 & $384^2$ & 46.9M & 32.2B & 84.3 \\
        & CaiT-S-36 & $384^2$ & 68M & 48.0B & 85.0 \\
        & CaiT-M-24 & $384^2$ & 185.9M & 116.1B & 84.5 \\
        \cmidrule(lr){2-2} \cmidrule(lr){3-3} \cmidrule(lr){4-5} \cmidrule(l){6-6}
        & DeepViT-S & $224^2$ & 27M & 6.2B &  82.3 \\
        & DeepViT-L & $224^2$ & 55M & 12.5B &  83.1 \\
        \cmidrule(r){1-2} \cmidrule(lr){3-3} \cmidrule(lr){4-5} \cmidrule(l){6-6}
        \multirow{7}{*}{Multi-Stage TFM Only} 
        & PVT-Small  & $224^2$ & 24.5M & 3.8B & 79.8 \\
        & PVT-Medium & $224^2$ & 44.2M & 6.7B & 81.2 \\
        & PVT-Large  & $224^2$ & 61.5M & 9.8B & 81.7 \\
        \cmidrule(lr){2-2} \cmidrule(lr){3-3} \cmidrule(lr){4-5} \cmidrule(l){6-6}
        & Swin-T  & $224^2$ & 29M  & 4.5B & 81.3 \\
        & Swin-S  & $224^2$ & 50M  & 8.7B & 83.0 \\
        & Swin-B  & $224^2$ & 88M  & 15.4B & 83.3 \\
        & Swin-B  & $384^2$ & 88M  & 47.0B & 84.2 \\
        \cmidrule(r){1-2} \cmidrule(lr){3-3} \cmidrule(lr){4-5} \cmidrule(l){6-6}
        \multirow{9}{*}{Multi-Stage Conv+TFM} 
        & BotNet-T7  & $384^2$ & 75.1M & 45.80B & 84.7 \\
        & LambdaResNet-420 & $320^2$ & - & - & 84.8 \\
        \cmidrule(lr){2-2} \cmidrule(lr){3-3} \cmidrule(lr){4-5} \cmidrule(l){6-6}
        & T2T-ViT-14 & $224^2$ & 21.5M &  6.1B & 81.7 \\
        & T2T-ViT-19 & $224^2$ & 39.2M &  9.8B & 82.2 \\
        & T2T-ViT-24 & $224^2$ & 64.1M & 15.0B & 82.6 \\
        \cmidrule(lr){2-2} \cmidrule(lr){3-3} \cmidrule(lr){4-5} \cmidrule(l){6-6}
        & CvT-13   & $224^2$ & 20M &  4.5B  & 81.6 \\
        & CvT-21   & $224^2$ & 32M &  7.1B  & 82.5 \\
        & CvT-13   & $384^2$ & 20M &  16.3B & 83.0 \\
        & CvT-21   & $384^2$ & 32M &  24.9B & 83.3 \\
        \cmidrule(r){1-2} \cmidrule(lr){3-3} \cmidrule(lr){4-5} \cmidrule(l){6-6}
        \multirow{10}{*}{\makecell{Proposed \\ Multi-Stage Conv+TFM}} 
        & \name-0 & $224^2$ & 25M  & 4.2B & 81.6 \\
        & \name-1 & $224^2$ & 42M  & 8.4B & 83.3 \\
        & \name-2 & $224^2$ & 75M  & 15.7B & 84.1 \\
        & \name-3 & $224^2$ & 168M & 34.7B & 84.5 \\
        \cmidrule(lr){2-2} \cmidrule(lr){3-3} \cmidrule(lr){4-5} \cmidrule(l){6-6}
        & \name-0 & $384^2$ & 25M  & 13.4B & 83.9 \\
        & \name-1 & $384^2$ & 42M  & 27.4B & 85.1 \\
        & \name-2 & $384^2$ & 75M  & 49.8B & 85.7 \\
        & \name-3 & $384^2$ & 168M & 107.4B & 85.8 \\
        & \name-2 & $512^2$ & 75M  & 96.7B & 85.9 \\
        & \name-3 & $512^2$ & 168M & 203.1B  & \bf 86.0 \\
        \bottomrule
    \end{tabular}
    \label{tab:i1k_result}
\end{table}

\begin{table}[!ht]
\footnotesize
    \centering
    \caption{Complete comparison under the ImageNet-21K pre-training + ImageNet-1K finetuning set up. ``PT-RA'' denotes applying RandAugment during 21K pre-training and ``E150'' means 150 epochs of pre-training, which is longer than the standard 90 epochs.}
    \begin{tabular}{c l c c c c}
        \toprule
        \multicolumn{2}{c}{\bf Models} & \bf Eval Size & \bf \#Params & \bf \#FLOPs & \bf Top-1 Accuracy \\
        \midrule
        \multirow{3}{*}{Conv Only}
        & ENetV2-S & $384^2$ & 24M  & 8.8B & 85.0 \\
        & ENetV2-M & $480^2$ & 55M  & 24B  & 86.1 \\
        & ENetV2-L & $480^2$ & 121M & 53B  & 86.8 \\
        \cmidrule(r){1-2} \cmidrule(lr){3-3} \cmidrule(lr){4-5} \cmidrule(l){6-6}
        \multirow{2}{*}{ViT-Stem TFM Only}
        & ViT-B/16 & $384^2$ & 87M  & 55.4B  & 84.6 \\
        & ViT-L/16 & $384^2$ & 304M & 190.7B & 85.3 \\
        \cmidrule(r){1-2} \cmidrule(lr){3-3} \cmidrule(lr){4-5} \cmidrule(l){6-6}
        \multirow{4}{*}{Multi-Stage TFM Only} 
        & HaloNet-H4 & $384^2$ & 85M & - & 85.6 \\
        & HaloNet-H4 & $512^2$ & 85M & - & 85.8 \\
        \cmidrule(r){2-2} \cmidrule(lr){3-3} \cmidrule(lr){4-5} \cmidrule(l){6-6}
        & Swin-B  & $384^2$ & 88M  & 47.0B  & 86.0 \\
        & Swin-L  & $384^2$ & 197M & 103.9B & 86.4 \\
        \cmidrule(r){1-2} \cmidrule(lr){3-3} \cmidrule(lr){4-5} \cmidrule(l){6-6}
        \multirow{3}{*}{Multi-Stage Conv+TFM}
        & HaloNet-Conv-H4 & $384^2$ & 87M & - & 85.5 \\
        & HaloNet-Conv-H4 & $512^2$ & 87M & - & 85.8 \\
        \cmidrule(r){1-2} \cmidrule(lr){3-3} \cmidrule(lr){4-5} \cmidrule(l){6-6}
        & CvT-13  & $384^2$ & 20M  & 16B    & 83.3 \\
        & CvT-21  & $384^2$ & 32M  & 25B    & 84.9 \\
        & CvT-W24 & $384^2$ & 277M & 193.2B & 87.7 \\
        \cmidrule(r){1-2} \cmidrule(lr){3-3} \cmidrule(lr){4-5} \cmidrule(l){6-6}
        \multirow{8}{*}{\makecell{Proposed \\ Multi-Stage Conv+TFM}}
        & \name-2 & $384^2$ & 75M  & 49.8B & 87.1 \\
        & \name-3 & $384^2$ & 168M & 107.4B & 87.6 \\
        & \name-4 & $384^2$ & 275M & 189.5B & 87.9 \\
        & \quad + PT-RA & $384^2$ & 275M & 189.5B & 88.3 \\
        & \quad + PT-RA-E150 & $384^2$ & 275M & 189.5B & 88.4 \\
        \cmidrule(r){2-2} \cmidrule(lr){3-3} \cmidrule(lr){4-5} \cmidrule(l){6-6}
        & \name-2 & $512^2$ & 75M  & 96.7B & 87.3 \\
        & \name-3 & $512^2$ & 168M & 203.1B  & 87.9 \\
        & \name-4 & $512^2$ & 275M & 360.9B & 88.1 \\
        & \quad + PT-RA & $512^2$ & 275M & 360.9B & 88.4 \\
        & \quad + PT-RA-E150 & $512^2$ & 275M & 360.9B & \bf 88.56 \\
        \bottomrule
    \end{tabular}
    \label{tab:21k_result}
\end{table}

\end{document}